\documentclass[10pt,journal,compsoc]{IEEEtran}
\ifCLASSOPTIONcompsoc
  \usepackage[nocompress]{cite}
\else
  \usepackage{cite}
\fi

\usepackage[utf8]{inputenc}
\usepackage{amsmath}
\usepackage{amsfonts}
\usepackage{graphicx}
\usepackage{multirow}
\usepackage{epsfig}
\usepackage{setspace}
\usepackage{graphicx}
\usepackage{caption}
\usepackage{color, algorithm, algpseudocode}
\ifCLASSINFOpdf
\else
\fi
\usepackage{url}

\begin{document}
%
\title{HAKE: A Knowledge Engine Foundation for Human Activity Understanding}
%
%
%
%

\author{Yong-Lu~Li, Xinpeng Liu, Xiaoqian Wu, Yizhuo Li, Zuoyu Qiu, Liang Xu, Yue Xu, Hao-Shu Fang, Cewu Lu
\IEEEcompsocitemizethanks{
\IEEEcompsocthanksitem Yong-Lu Li is with the Qing Yuan Research Institute at the Shanghai Jiao Tong University, Shanghai, China. E-mail: \{yonglu\_li\}@sjtu.edu.cn\protect\\
\IEEEcompsocthanksitem Xinpeng Liu, Xiaoqian Wu, Yizhuo Li, Zuoyu Qiu, Liang Xu, Yue Xu, and Hao-Shu Fang are with the Department of Computer Science and Engineering at the Shanghai Jiao Tong University, Shanghai, China. E-mail: \{enlighten, liyizhuo, 17803091056, liangxu, silicxuyue\}@sjtu.edu.cn, {xinpengliu, fhaoshu}0907@gmail.com.\protect\\
\IEEEcompsocthanksitem Cewu Lu is the corresponding author, a member of Shanghai Digital Medicine Innovation Center, Ruijin Hospital and MoE Key Lab of AI, AI Institute, Shanghai Jiao Tong University, China, and Shanghai Qi Zhi Institute. E-mail: lucewu@sjtu.edu.cn.
}}

\IEEEtitleabstractindextext{%
\begin{abstract}
Human activity understanding is of widespread interest in artificial intelligence and spans diverse applications like health care and behavior analysis. Although there have been advances in deep learning, it remains challenging. The object recognition-like solutions usually try to map pixels to semantics directly, but activity patterns are much different from object patterns, thus hindering success. In this work, we propose a novel paradigm to reformulate this task in two stages: first mapping pixels to an intermediate space spanned by atomic activity primitives, then programming detected primitives with interpretable logic rules to infer semantics. To afford a representative primitive space, we build a knowledge base including 26+ M primitive labels and logic rules from human priors or automatic discovering. Our framework, the \textit{\textbf{H}uman \textbf{A}ctivity \textbf{K}nowledge \textbf{E}ngine} (\textit{\textbf{HAKE}}), exhibits superior generalization ability and performance upon canonical methods on challenging benchmarks. Code and data are available at \url{http://hake-mvig.cn/}.
\end{abstract}


}

\maketitle

\IEEEdisplaynontitleabstractindextext

%
\IEEEpeerreviewmaketitle

\IEEEraisesectionheading{
\section{Introduction}
\label{sec:introduction}}

\begin{figure*}[!ht]
\begin{center}
    \includegraphics[width=0.95\textwidth]{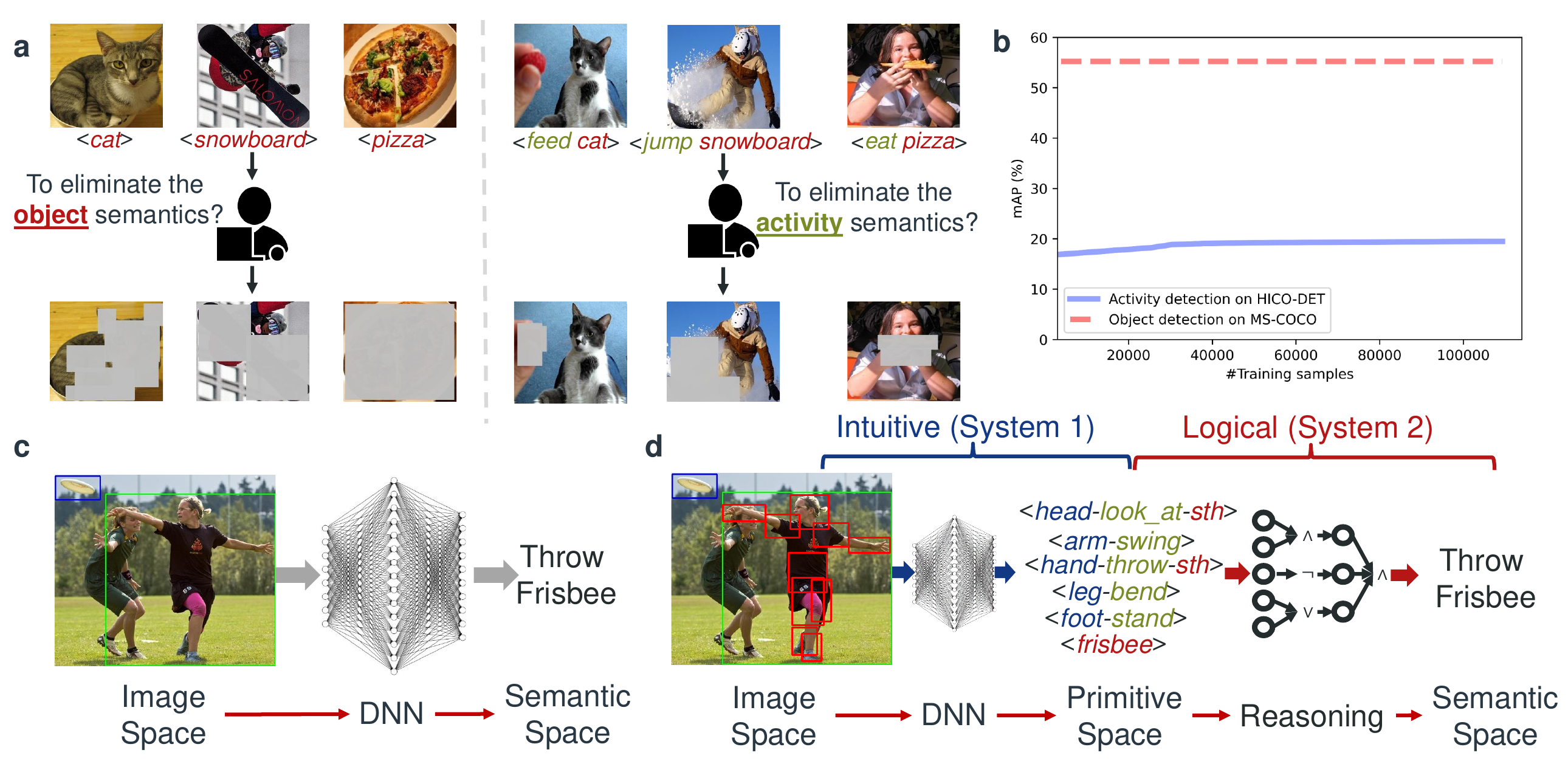}
\end{center}
    \vspace{-15px}
	\caption{Object Recognition vs. Activity Recognition. \textbf{a.} SCR test reveals the difference between activity and object visual patterns. \textbf{b.} Bottleneck of direct mapping. Given the same order of magnitude of training images, it rapidly saturates on the HICO-DET~\cite{hicodet} test set and performs much worse than object recognition on MS-COCO~\cite{coco} (dotted line, 55+ mAP (mean Average Precision, \%)). \textbf{c-d.} Direct mapping and our two-stage paradigm. We introduce an intermediate \textbf{primitive space} to embed activity information and infer semantics via \textbf{primitive programming}.}
	\label{Fig:1}
	\vspace{-15px}
\end{figure*}

\IEEEPARstart{V}{isual} activity understanding is a fundamental AI problem with great impacts, \textit{e.g.}, 
helping health care~\cite{feifeinature}, advancing robot skill learning from humans~\cite{smith2019avid}, \textit{etc}. 
All these require accurate activity recognition. However, current systems are far from qualified, even in this booming deep learning era. This is mainly because the conventional pattern recognition route, which attempts to learn a direct mapping from the image space to the activity semantic space, may not be suitable for activity recognition.

The underlying reason is that taking the most representative object recognition as a comparison case, the \textit{activity} pattern is much different from the \textit{object} pattern. Similar success cannot be achieved by direct mapping. To prove this, a novel measurement depicting image semantic ``density'' is introduced: \textbf{Semantic Coverage Rate (SCR)}.
In a bounding box containing an entity (person/object/interactive person-object pair, Fig.~\ref{Fig:1}a), the \textit{area proportion of key regions} carrying the indispensable semantics for recognition is SCR. The principle for measuring SCR is to ask human participants to mask the \textit{smallest} box regions to make other human participants unable to distinguish the entity (Suppl.~Sec.~1). 
For object recognition (Fig.~\ref{Fig:1}a), \textit{pizza’s} SCR is \textbf{93.7\%}. In contrast, more concentrated masks are usually made to activity images, leading to smaller SCRs: \textit{eat pizza} has an SCR of \textbf{15.3\%} (unmasked regions $\approx$ ``clutter''). 
By testing on 1,500 images, the average SCRs of object and activity images are \textbf{61.0\%} and \textbf{12.2\%}, respectively. 
Hence, we conclude that activity recognition is greatly different from object recognition. A lower SCR requires us to mine \textit{sparse} semantics from ``cluttered'' scenes, thus hindering direct mapping (Fig.~\ref{Fig:1}c) from achieving efficacy in object recognition, even with large-scale data. In Fig.~\ref{Fig:1}b, given a quantity of data, it can only achieve limited performance (detailed in Sec.~\ref{sec:bottleneck-test}), which is far from successful object recognition~\cite{coco}.

\begin{figure}[!ht]
\begin{center}
    \includegraphics[width=0.5\textwidth]{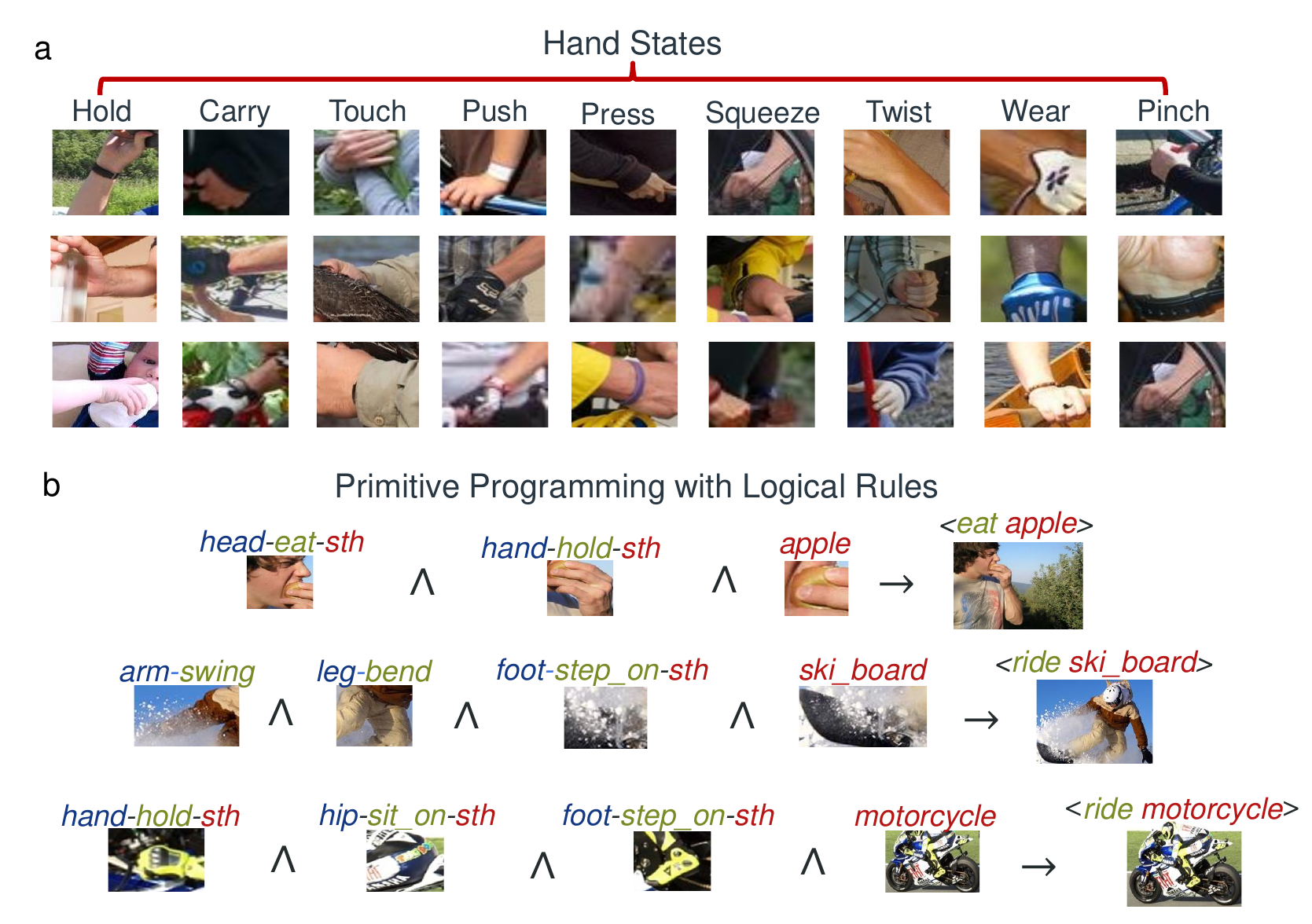}
\end{center}
    \vspace{-15px}
	\caption{Nature of activity perception: atomic primitive and logical reasoning. \textbf{a.} Primitives exist such as body part states~\cite{li2020pastanet} and objects. Here, we show some common hand states. \textbf{b.} Activities can be inferred by programming primitives following logical rules.}
	\label{Fig:2}
	\vspace{-0.5cm}
\end{figure}

To find a better solution, a natural method is to observe how humans recognize activities. An empirical study is conducted to ask participants to explain how they differentiate activities. 
First, local visual cues are frequently mentioned. For activities \textit{run} and \textit{walk}, the claimed differences mainly lie in the \textit{moving state of legs}, but the upper body affords limited cues, which is in line with the discovery of a smaller SCR. For \textit{drink}, the main criterion is often the existence of \textit{hand-hold-sth}, \textit{head-contact\_with-sth}, \textit{cup}. These \textbf{atomic} and \textbf{shareable} cues can be seen as \textbf{primitives} (Fig.~\ref{Fig:2}a). We believe that humans can unconsciously discover primitives (like System 1~\cite{bengio-system}). Moreover, cognitive neuroscience studies also found that certain brain regions in the ventral stream (EBA) process visual information about body parts and objects~\cite{zimmermann2018extrastriate}. 
Second, primitives can be seen as the \textbf{causes} of activities: participants can logically combine \textit{hand-hold-sth}, \textit{head-contact\_with-sth}, \textit{cup} to infer \textit{drink} (like System 2~\cite{bengio-system}). This is in accordance with the cognition study that humans combine basic components to perceive concepts~\cite{hoffman1983parts,biederman1987recognition,lake2017building}. 
We argue that activity perception rests on the ability to \textit{discover primitives}, while the efficacy of human inference arises from \textit{logical reasoning} to program primitives into semantics with compositional generalization (Fig.~\ref{Fig:2}b). 
Thus, we propose \textbf{H}uman \textbf{A}ctivity \textbf{K}nowledge \textbf{E}ngine (\textbf{HAKE}), which is an end-to-end differentiable two-stage system (Fig.~\ref{Fig:1}d):

(1) \textbf{Intermediate primitive space} embeds activity information in images with limited and representative primitives. We build a comprehensive knowledge base by crowd-sourcing. As primitive space is sparse~\cite{li2020pastanet}, we can cover most primitives from daily activities, \textit{i.e.}, \textit{one-time labeling} and \textit{transferability}. HAKE trained on our knowledge base can detect primitives well and focus only on primitives instead of the whole image, thus achieving an \textit{equivalent higher} SCR. 
    
(2) \textbf{A reasoning engine} programs detected primitives into semantics with \textit{explicit} logic rules and updates the rules during reasoning. That is, diverse activities can be composed of a finite set of primitives via logical reasoning with \textit{compositional generalization}.

\section{Related Works}
\noindent{\bf Activity Understanding.}
Benefited by deep learning and large-scale datasets, activity understanding has achieved huge improvements recently. There are mainly image-based~\cite{hico, vcoco}, video-based~\cite{ava, Kinetics, UCF101, JhuangICCV2013} and skeleton-based~\cite{Vemulapalli_2014_CVPR, Du_2015_CVPR} methods. 
Human activities have a hierarchical structure and include diverse verbs, so it is hard to define an explicit organization for their categories. Existing datasets~\cite{ava, hico, activitynet, vcoco} often have a large difference in definition, thus transferring knowledge from one dataset to another is ineffective. 
Plenty of works based on CNN, 3D-CNN, GNN, or GCN have been proposed to address activity understanding~\cite{Delaitre2011Learning, Gkioxari_2015_ICCV, Du_2015_CVPR, NIPS2014_5353, Feichtenhofer_2016_CVPR, 6165309, Tran_2015_ICCV, Sun_2015_ICCV}. 
But compared with object detection~\cite{faster} or pose estimation~\cite{fang2017rmpe}, its performance is still limited. 

\noindent{\bf Human-Object Interaction (HOI) Learning.}
HOI consists of the most important part of daily activities. As a sub-filed of visual relationship learning, HOI understanding has attracted a lot of attention and several large datasets~\cite{hicodet,vcoco} have been released.
Meanwhile, to prompt this field, many deep learning based methods~\cite{Gkioxari2017Detecting,gao2018ican,interactiveness,gpnn,analogy,vcl} were proposed.
Wu~et.~al.~\cite{hicodet} proposed a multi-stream framework followed by subsequent works~\cite{gao2018ican,interactiveness,vcl}.
GPNN~\cite{gpnn} used a graphical model to address HOI detection. 
iCAN~\cite{gao2018ican} adopted self-attention to correlate the human, object, and context. 
TIN~\cite{interactiveness} modeled interactiveness to suppress non-interactive pairs. 
VCL~\cite{vcl} exploited the compositional characteristic of HOI. 
Also, some works~\cite{analogy,zhong2020polysemy} dug into the relationship among HOIs, while IDN~\cite{idn} focused on how HOIs are constructed via integration and decomposition.
Moreover, there also appeared one-stage methods~\cite{ppdm} that directly detected HOI triplets. 
Besides the works based on convolutional neural networks (CNN), recently Transformer-based methods~\cite{qpic} are proposed and achieved decent improvements. 

\noindent{\bf Body Part based Activity Understanding.}
Usually, previous works used instance-level patterns as the cue. 
However, some approaches were studied to utilize finer-grained human body part pattern~\cite{Du_2015_CVPR,zhao2017single,Fang2018Pairwise} for better understanding. 
Khan~et.~al.~\cite{khan2014semantic} combines the features of instance and parts to operate the recognition.
Gkioxari~et.~al.~\cite{Gkioxari_2015_ICCV} detects both the instance and parts and inputs them all into a classifier.
Yao~et.~al.~\cite{Yao2010Modeling} builds a graphical model and embeds parts appearance as nodes, and uses them with object feature and pose to predict the HOIs.
Previous works mainly utilized the part \textit{appearance}/\textit{location}, but few studies tried to divide the instance actions into discrete part-level semantic tokens and refer to them as the basic components of activity concepts. 
In comparison, we aim at building human part semantics as the \textbf{reusable} and \textbf{transferable} primitives. Furthermore, a neural-symbolic reasoning system is built to program primitives into activity concepts.

\noindent{\bf Neural-Symbolic Reasoning.} 
The integration of neural models with logic-based symbolic models provides an AI system capable of bridging lower-level information processing (for perception and pattern recognition) and higher-level abstract knowledge (for reasoning and explanation)~\cite{garcez2019neural}. 
In neural-symbolic computing, knowledge is represented in symbolic form, whereas a neural network computes learning and reasoning.
Recently, researchers combined neural-symbolic reasoning with visual question answering (VQA) task, and proposed methods including neural module networks~\cite{andreas2016neural}, neural state machine~\cite{hudson2019learning}, symbolic program execution on structural scene representation~\cite{yi2018neural}, neuro-symbolic concept learner~\cite{mao2019neuro}. Besides, \cite{dong2019neural} proposed neural logic machines for relational reasoning and decision-making tasks.
Besides, there are also efforts adopting syntactic approaches on different tasks, including visual grounding~\cite{shi2019visually}, motion understanding~\cite{kulal2021hierarchical} and scene understanding~\cite{liu2019learning}.
Inspired by this line of methods, we use Deep Neural Networks (DNN) to represent the activity primitives as symbols meanwhile borrowing the power of symbolic reasoning to use logic rules to program these primitives.

\section{Bottleneck of Canonical Direct Mapping}
\label{sec:bottleneck-test}
First, we conduct an experiment to demonstrate that canonical direct mapping suffers from severe performance bottleneck problems. We choose a canonical action detection method TIN~\cite{interactiveness} as a representation of the direct mapping methods. Then, we train it with a different number of images from HAKE data and evaluate its performance on the HICO-DET~\cite{hicodet} test set. 
For each run, the TIN model is trained for 30 epochs using an SGD optimizer with a learning rate of 0.001 and a momentum of 0.9. The ratio of positive and negative samples is 1:4. A late fusion strategy is adopted. For HICO-DET, we adopt the Default mAP metric as proposed in HICO-DET~\cite{hicodet}.

The result is illustrated in Fig.~\ref{Fig:1}b. It shows that, in the beginning, the increasing amount of data indeed helps boost the performance of the direct mapping method in a nearly linear manner. However, as more and more data get involved, the performance gain becomes less significant and rapidly saturates. 
This indicates that it is hard for canonical direct mapping methods to make the best use of the increasing data without the help of knowledge and reasoning. Even providing more labeled data, the direct mapping may not achieve the success of object detection on MS-COCO~\cite{coco} (CNN-based, 55+ mAP) given the same magnitude ($10^5$) of training images. Besides, activity images are also much more difficult to annotate than object images, considering the complex patterns and ambiguities of human activities.

\section{Method}
\label{sec:hake-pipeline}

\subsection{Overview}
\begin{figure*}[!ht]
\begin{center}
    \includegraphics[width=0.95\textwidth]{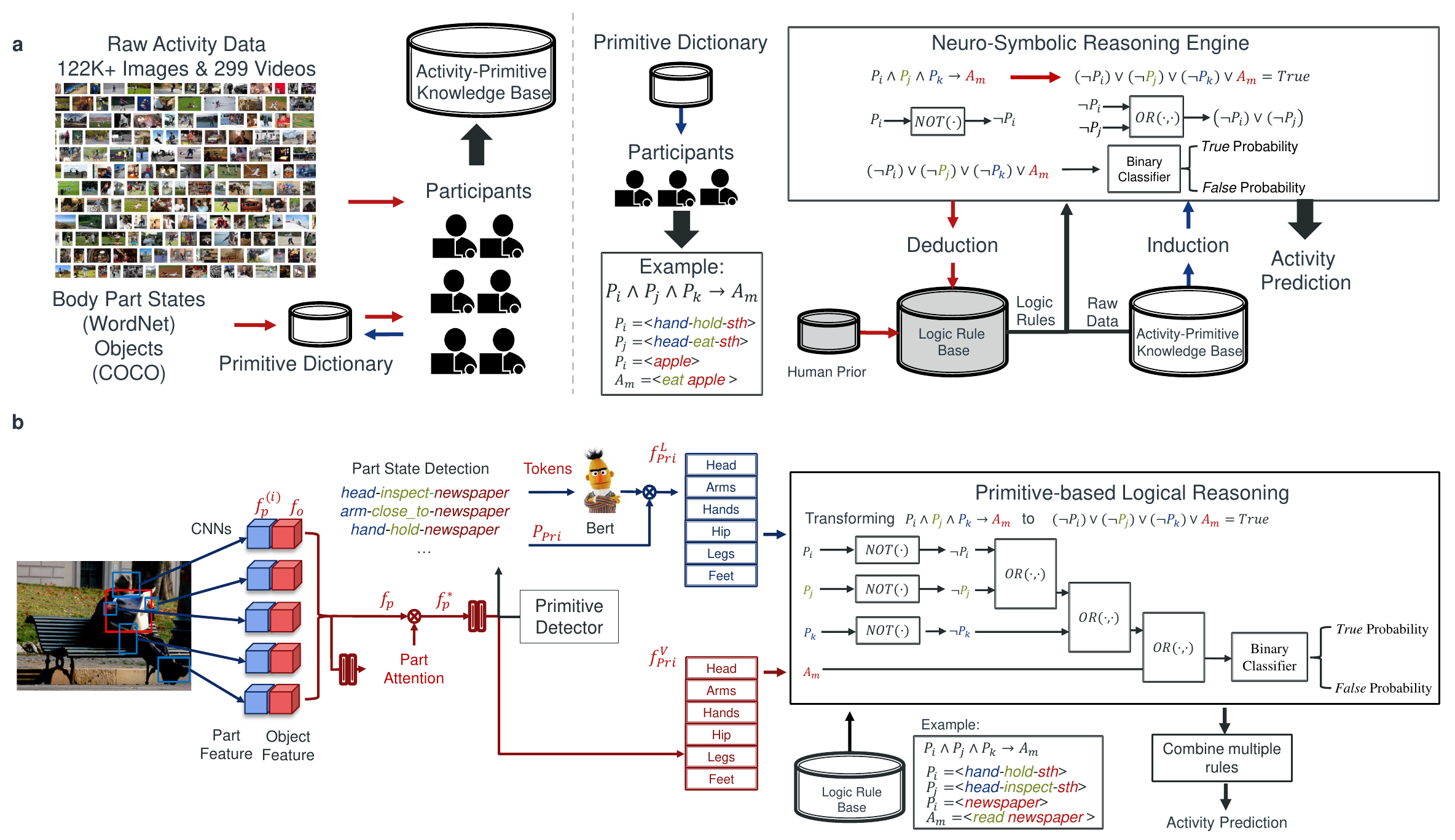}
\end{center}
    \vspace{-15px}
	\caption{HAKE Overview. \textbf{a.} We cast activity understanding into: 
	\textbf{a(1)} Knowledge base construction: annotating large-scale activity-primitive labels to afford accurate primitive detection. 
	\textbf{a(2)} Reasoning: Given detected primitives, adopting neuro-symbolic reasoning to program them into semantics. 
	\textbf{b.} Detailed pipeline. 
	\textbf{b(1)} Primitive detection and Activity2Vec. Given an image, we utilize the detectors~\cite{faster,fang2017rmpe} to locate the human/object and human body parts. Then we use a simple CNN model together with Bert~\cite{devlin2018bert} to extract the visual and linguistic features of primitives via primitive detection. 
	\textbf{b(2)} Primitive-based logical reasoning. With the two kinds of representations from Activity2Vec, we operate logical reasoning in a neuro-symbolic paradigm following the prior and auto-discovered logic rules. Here, $NOT(\cdot)$ and $OR(\cdot,\cdot)$ modules are shared by all events but \textbf{drawn separately} here for clarity.} 
	\label{Fig:3}
	\vspace{-0.5cm}
\end{figure*}

As depicted in Fig.~\ref{Fig:3}, HAKE casts activity understanding into two sub-problems: (a) \textbf{knowledge base building}, (b) \textbf{neuro-symbolic reasoning}. 

\subsubsection{Building Human Activity Knowledge Base}
Firstly, HAKE should detect primitives ``unconsciously'' like System 1~\cite{bengio-system}. 
Thus, we built a knowledge base including abundant activity-primitive labels. 
To define and discover primitives, we conduct a beforehand user study: given activity images/videos, participants should give decisive primitives for activities from an \textbf{initial primitive dictionary} including 200 human body Part States (PaSta~\cite{li2020pastanet}), 80 common objects~\cite{coco}, and 400 scenes~\cite{zhou2017places}. 
Each primitive is depicted as a phrase token, \textit{e.g.}, \textit{hand-touch-sth}, \textit{chair}, \textit{classroom}. This dictionary grows as the annotation progresses, and participants can supplement primitives to better explain their decisions. 
After annotating 122 K+ images and 234 K+ video frames~\cite{ava} covering 156 activities, we found that:
1) most participants believed body parts were the foremost activity semantic carriers. 
2) PaSta classes were limited. After exhaustively checking 26.2 M+ manually labeled PaSta, only approximately 100 classes were prominent. 
3) Object sometimes makes a small difference, \textit{e.g.}, in \textit{inspect-sth, touch-sth}. 
4) Few participants believed that scene always matters, \textit{e.g.}, though rarely, we can \textit{play football} in \textit{living room}. 

In light of these, our dictionary would contain as many cues as possible and leave the choice of use to reasoning policy. Though scalable, it already contains enough primitives to compose common activities, so we usually do not need to supplement primitives for new activities. 
In total, HAKE includes \textbf{357 K+} images/frames, \textbf{673 K+} persons, \textbf{220 K+} object primitives, and \textbf{26.4 M+} \textit{PaSta} primitives. With abundant annotations, a simple CNN-based detector detects primitives well and achieves \textbf{42.2} mAP (Sec.~\ref{sec:primitive_result}).

\subsubsection{Constructing Logic Rule Base and Reasoning Engine}
Second, to determine activities, a causal protocol consisting of a \textbf{logic rule base} and a \textbf{reasoning engine} is proposed. After detecting primitives, we use a DNN to extract visual and linguistic representations to represent primitives $P=\{P_i\}_{i=1}^N$ and activities $A=\{A_m\}_{m=1}^M$. 
As interpretable symbolic reasoning can capture causal primitive-activity relations, we leverage it to program primitives following logic rules. A logic rule base is initialized to import common sense: participants are asked to describe the \textbf{causes} (primitives) of \textbf{effects} (activities). Each activity has \textit{initial} multi-rule from different participants to ensure diversity. 
For example (Fig.~\ref{Fig:3}b), $P_i, P_j, P_k$ represent \textit{head-read-sth}, \textit{hand-hold-sth}, \textit{newspaper} and $A_m$ indicates \textit{read newspaper}, a rule is expressed as $P_i\land P_j\land P_k\rightarrow A_m$ ($\land$: AND, $\rightarrow$: implication). $P_i,P_j,P_k,A_m$ are seen as \textbf{events} that are occurring/True or not/False. When $P_i,P_j,P_k$ are True simultaneously, $A_m$ is True. 
For simplicity, we turn $\rightarrow, \land$ into $\vee, \lnot$ via $x\rightarrow y \Leftrightarrow\lnot x\ \vee y$. $\lnot$ and $\vee$ are implemented as functions $NOT(\cdot)$, \ $OR(\cdot,\cdot)$ with Multi-Layer Perceptrons (MLPs) that are reusable for all events. 
We set logic laws (idempotence, complementation, \textit{etc}.) as optimized objectives~\cite{shi2020neural} imposed on all events to attain logical operations via backpropagation. Then, the expression output is fed into a discriminator to estimate the true probability of an event. 

Given a sample, multi-rule predictions of all activities are generated concurrently. We use voting to combine multi-decision into the final prediction via multi-head attention~\cite{attention}. 
Besides, to better capture the causal relations, we propose an \textbf{inductive-deductive policy} to make the rule base scalable instead of using rule templates or enumerating possible rules~\cite{rocktaschel2017end}.
Although annotating rules is much easier than annotating images/videos, we can discover novel rules with the initial prior rules (Sec.~\ref{sec:rule_up_eval}): 
1) Inductively concluding rules from observations. As activity-primitive annotations can be seen as rule instantiations, we randomly select them as rule candidates or heuristically generate candidates according to the human prior rule distribution. 
2) Deductively evaluating rule candidates in practical training. 
With the updating reasoning model, good candidates inducing minor losses are selected as they are more compatible and general with various contexts.
They are updated into the rule base and in turn boost the performance. 
The bidirectional process can robustly discover rules and is relatively insensitive to the annotation shortage, as we can sample new rules according to the human prior distribution.
Finally, we discover \textbf{4,090} rules for 156 activities. 
From the experiments, we find that HAKE requires only several rules for each activity to perform well.

Next, we detail our framework in Sec.~\ref{sec:primitive_det}-\ref{sec:unified_infer}.

\subsection{Primitive Detection} 
\label{sec:primitive_det}
Given an image, detected body part boxes, and object boxes $\mathcal{I}, b_p, b_o$ (with object classification), we operate primitive detection following Fig.~\ref{Fig:3}b.
We assume the number of body parts is $m$. 
In detail, we first get the \textbf{part} features $f_p = \{f^{(i)}_p\}_{i=1}^{m}$ and \textbf{object} feature $f_o$ via ROI pooling from a feature extractor $\mathcal{F}$ (Faster R-CNN~\cite{faster} pre-trained on COCO~\cite{coco} for image and 3D convolution backbone~\cite{Kinetics} pre-trained on Kinetics~\cite{Kinetics} for video).
For body-only motions~\cite{ava} (\textit{e.g.}, \textit{walk}, \textit{run}) that are not involved with objects, $b_o$ is replaced with the whole image coordinates. That is, we input the whole image feature $f_c$ as $f_o$. Besides, $f_c$ can also represent the \textbf{scene} primitive.
Then, we use these features to predict the \textbf{part relevance} indicating the contribution of a body part for recognizing an activity.  
For example, feet usually have weak correlations with \textit{drink with cup}. And in \textit{eat apple}, only hands and head are essential.
For each body part, we concatenate the part feature $f^{(i)}_p$ from $b^{(i)}_p$ and object feature $f_o$ from $b_o$ as the inputs. 
All features will be input to a part relevance predictor $\mathcal{R}(\cdot)$, which contains FC layers and Sigmoids, getting $a=\{a_{i}\}^{m}_{i=1}$ ($a_{i} \in [0,1]$) following
\begin{eqnarray}
    \label{eq:pa-recog}
    a & = & \mathcal{R}(f_p, f_o).
\end{eqnarray}
The relevance/attention labels can be \textbf{converted} from primitive labels directly, \textit{i.e.} the attention label will be \textit{one} unless its primitive label is \textit{no\_activity}, which means this part contributes nothing to the inference.
Next, we perform primitive classification. 
For each part, we concatenate $f^{(i)}_p$ with $f_o$ and input them into a max pooling and FC layers, thus obtaining the primitive score $\mathcal{S}_{P}^{(i)}$ for the $i$-th part:
\begin{eqnarray}
    \label{eq:pasta-recog}
    \mathcal{S}^{(i)}_{P} & = & \mathcal{P}_{P}^{(i)}(f^{(i)}_p, f_o).
\end{eqnarray}
Because a part can have multiple states, \textit{e.g.} \textit{head} performs \textit{eat} and \textit{watch} simultaneously, we use multiple Sigmoids to operate the multi-label classification.
For part relevance prediction and primitive classification, we construct cross-entropy losses ${L}^{(i)}_{a}$ and ${L}^{(i)}_{P}$ for each part.
Formally, for a pair (HOI) consisting of a person and an interacted object or a single person (body-only motion), the final loss $\mathcal{L}_{P}$ is
\begin{eqnarray}
    \label{eq:part-loss}
    \mathcal{L}_{P} & = & \sum^{m}_{i} (\mathcal{L}^{(i)}_{P} + \mathcal{L}^{(i)}_{a}).
\end{eqnarray}

\subsection{Activity2Vec}
\label{sec:a2v}
As mentioned before, we define the primitives according to the most common activities via crowdsourcing. That is, choosing the part-level verbs that are most often used to compose and describe the activities by a large number of annotators. Therefore, primitives can be seen as the fundamental components of activities. Meanwhile, primitive detection performs well. 
Thus, we can utilize primitive detection based on HAKE to learn primitive representation with good transferability. They can be used as ``symbols'' to reason out the ongoing activities in both supervised and transfer learning. Here, the primitive representation extractor is called as Activity2Vec.
Under such circumstances, HAKE works like ImageNet~\cite{imagenet} as the knowledge base. The HAKE pre-trained Activity2Vec functions as a knowledge engine to transform low-level activity patterns to the primitive space for subsequent primitive programming.

\noindent{\bf Visual Primitive Representation.} 
First, we extract visual primitive representations from primitive detection.
We extract the feature from the last FC layer in $\{\mathcal{P}_{P}^{(i)}(\cdot)\}_{i=1}^{m}$ as the raw primitive visual representation $\{f^{V(i)}_{P}\}_{i=1}^{m}$ for each part. 
The predicted part relevance $a$ provides a cue on how important a part is to the activity. To utilize this cue, we further use $a$ to \textit{re-weight} the raw primitive visual representation to generate the primitive visual representation $f^{V}_{P}$:
\begin{eqnarray}
    \label{eq:vis-pasta}
    f^{V}_{P} = \{f^{V(i)}_{P} * a_i\}_{i=1}^{m}.
\end{eqnarray}

\noindent{\bf Linguistic Primitive Representation.} 
To further enhance the representation ability, we utilize the uncased BERT-Base pre-trained model~\cite{devlin2018bert} to represent primitives as event vectors. 
Language priors are useful in visual concept understanding~\cite{Lu2016Visual,vinyals2015show}. 
Thus the combination of visual and language knowledge is a good choice for establishing this mapping.
Bert~\cite{devlin2018bert} is a language understanding model that considers the context of words and uses a deep bidirectional transformer to extract contextual representations. It is trained with large-scale corpus databases such as Wikipedia, hence the generated embedding contains helpful implicit semantic knowledge about the activity and primitive.
For example, the description of the entry \textit{basketball} in Wikipedia: ``drag one's \textit{foot} without \textit{dribbling} the ball, to \textit{carry} it, or to \textit{hold} the ball with both \textit{hands}...\textit{placing} his \textit{hand} on the bottom of the ball;..known as \textit{carrying} the ball''.
In specific, if there are $n$ primitives in total, we reform each primitive into tokens $\{t^{j}_p, t^{j}_v, t^{j}_o\}^{n}_{j=1}$, \textit{e.g.}, $\langle part\_class, verb\_class, object\_class\rangle$, where $\langle object\_class \rangle$ comes from object detection.
Then we get primitive linguistic representation $f^L_{P}$ as:
\begin{eqnarray}
    \label{eq:bert-pasta}
    f^L_{P} & = & \{Bert(t^{k}_p, t^{k}_v, t^{k}_o)\}_{k=1}^n.
\end{eqnarray}
Also each activity is converted to $f^k_{A}$ following the above fashion, where ``part\_class'' is replaced by ``human'' as: 
\begin{eqnarray}
    \label{eq:bert-act}
    f_{A} & = & \{Bert('human', t^{k}_v, t^{k}_o)\}_{k=1}^n.
\end{eqnarray}
For body-only motions, we use $\langle part\_class, verb\_class\rangle$. We can also add $\langle scene\_class\rangle$ if necessary.

\noindent{\bf Primitive-Based Logical Reasoning.} 
We utilize the primitive representations from Activity2Vec as symbols to implement logical reasoning.
Both linguistic and visual representations are used.
Firstly, all representations $f_{P}^L, f_{P}^V, f_{A}$ are mapped into a lower-dimensional event space with MLPs, generating $e_{P}^V, e_{P}^L, e_{A}$.
The dimension reduction can accelerate the learning process without too many losses in performance.
As mentioned before, we turn all $\rightarrow$ into $\vee$ and $\neg$ via $x \rightarrow y \Leftrightarrow \neg x \vee y$.
Hence, logical expression like $P_i \wedge P_j \wedge P_k \rightarrow A_m$ is converted to $(\neg P_i) \vee (\neg P_j) \vee (\neg P_k) \vee A_m$. Then we only need logical operations $\lnot$ and $\lor$ to construct all logic rules to reduce the learning difficulty of logic operation.
Each time $OR(\cdot,\cdot)$ takes two compressed event vectors as the inputs and generates one output event vector representing the conjunction of two input events. This output vector can then be input to $OR(\cdot,\cdot)$ again together with another event vector to operate the subsequent conjunction. 
$NOT(\cdot)$ takes one compressed event vector each time and outputs its opposite event vector. 
In practice, we use \textit{probability} to measure the existence of \textit{primitive} events. 
$(\lnot P_i)\vee(\lnot P_j)\vee(\lnot P_k)\vee A_m$ tends to be true if and only if $P_i,P_j,P_k$ all have high probabilities of occurrence. 
As $e_{P}^V$ is naturally \textbf{probabilistic} extracted from DNN, we directly use it in logic reasoning.
As for $e_{P}^L$ transformed from the (happening) event \textit{truths} like $\langle hand, hold, cup\rangle$ that is \textbf{deterministic}, we cannot directly use it. 
Naturally, the \textbf{expectation} of linguistic representation is appropriate. 
For a primitive event $P_i$ represented as $e_{P}^L$ with probability $\mathcal{S}_{P}$ (from the Activity2Vec), its opposite event is $\lnot P_i$ represented as $NOT(e_{P}^L)$ with probability $1-\mathcal{S}_{P}$.
Thus the expectation is 
\begin{eqnarray}
    \label{eq:prob_comb}
    e_{P}^{L'} & = & e_{P}^L \mathcal{S}_{P} + NOT(e_{P}^L) (1-\mathcal{S}_{P}).
\end{eqnarray}
Besides, for visual representation, $e_{P}^{V'} = e_{P}^V$.
Hereinafter, we use $e_{P}^{'}$ to refer to $e_{P}^{L'}$ or $e_{P}^{V'}$ for clarity.
As for the target activity $A_m$, we use the linguistic representation truth $f_{A}$ (like $\langle human, eat, apple\rangle$) from Eq.~\ref{eq:bert-act} in logic reasoning.

For the $m$-th action category, we get its rule set as $R_m=\{r_i\}_{i=1}^{l_m}$, where $l_m$ is the rule number. 
For the $i$-th rule $r_i \in R_m$, assume its expression as: $P_a \wedge P_b \rightarrow A_m$, where $P_a, P_b$ is the $a$-th, $b$-th primitive, and $A_m$ is the $m$-th activity. 
For example, \textit{hand-catch-sth} $\wedge$ \textit{ball} $\rightarrow$ \textit{human-play-ball}.
Then we transform the expression into $(\neg P_a) \vee (\neg P_b) \vee A_m$, and calculate the vote vector via
\begin{equation}
    \label{eq:get_vote}
    e_{m}^{(i)} = OR(OR(NOT(e^{'}_{P_a}), NOT(e^{'}_{P_b})), e_{{A}_m}). 
\end{equation}
The vector $e_{m}^{(i)}$ implies the result of logical reasoning based on the $i$-th rule. All $OR(\cdot,\cdot), NOT(\cdot)$ are \textbf{shared} functions but written separately for clarity.
 
We explore two different ways to synthesize the rules: early combination and late combination.
For early combination, we get a new embedding $e_{m}^{(i)'}$ for each rule via
\begin{equation}
     \label{eq:mh-att}
    e_{m}^{(i)'} = \sum_{j=1}^{l_m} att_j^{(i)} * e_{m}^{(j)'},
\end{equation}
where the attention scores are obtained by Multi-Head Attention strategy~\cite{attention}. 
Then we concatenate all $e_{m}^{(i)'}$ and map it into an aggregated event vector $e_{m}$ via an MLP, and get action inference score $\mathcal{S}_{m}$ with a binary discriminator transforming $e_{m}$ to probability.
For the late combination, the process is the opposite. 
We first get the action prediction $\mathcal{S}_{m}^{(i)}$ from $e_{m}^{(i)}$.
Since different persons may have different understandings of activities, we synthesize the inference from different rules and average $\mathcal{S}_{m}^{(i)}$ to get the final result $\mathcal{S}_{m}$.
These two policies are used differently. In rule discovery and updating (Sec.~\ref{sec:rule_up_eval}), we utilize the late combination. Then, we freeze the updated rule base and finetune the whole reasoning module via early combination. 
We give more details in Suppl.~Sec.~3.3.

To perform logical operations, logical modules $NOT(\cdot)$ and $OR(\cdot, \cdot)$ should satisfy the basic laws of logic:
1) NOT negation: $\neg$ TRUE = FALSE, 2) NOT double negation: $\neg \neg x = x$, 3) OR idempotence: $x\vee x = x$, 4) OR complementation: $x\vee (\neg x) =$ TRUE, where $x \in \mathcal{X}$ refers to the event vector, and $\mathcal{X}$ is the event vector space.
We utilize the laws as the regularization objective via
\begin{equation}
\begin{aligned}
    \label{eq:gamma_i}
    \mathcal{L}_{reg} = \sum_{i=1}^4 \sum\limits_{x \in \mathcal{X}}  (\mathcal{S}_{ia}- \mathcal{S}_{ib})^2,
\end{aligned}
\end{equation}
where $i$ indicates the $i$-th logic law. In detail, $\mathcal{S}_{ia}, \mathcal{S}_{ib}$ are
\begin{eqnarray}
    \label{eq:gamma_i_append}
    \mathcal{S}_{1a} = \mathcal{J}(NOT(x)),
    & & \mathcal{S}_{1b} = 1-\mathcal{J}(x); \notag \\
    \mathcal{S}_{2a} = \mathcal{J}(NOT(NOT(x))),
    & & \mathcal{S}_{2b} = \mathcal{J}(x); \notag \\
    \mathcal{S}_{3a} = \mathcal{J}(OR(x,x)),
    & & \mathcal{S}_{3b} = \mathcal{J}(x); \notag \\
    \mathcal{S}_{4a} =  \mathcal{J}(OR(x,NOT(x))),
    & & \mathcal{S}_{4b} = 1.
\end{eqnarray}
Since the logic laws are universal despite input and output, we impose regularizations on \textit{all} event representations $x \in \mathcal{X}$ related to logical modules, including primitive, intermediate, and final event vectors.

Therefore, the loss of reasoning consists of two parts. 
First, the modules must follow the logical laws, so we use the logical regularizer loss $\mathcal{L}_{reg}$ for all event vectors.
Second, the modules should be able to perform accurate reasoning, so the classification loss $\mathcal{L}_{cls}$ is essential. The final loss of the logic reasoning module $\mathcal{L}_{LR}$ is constructed as
\begin{eqnarray}
    \label{eq:reason_loss}
    \mathcal{L}_{LR}=\alpha * \mathcal{L}_{reg}+\mathcal{L}_{cls},
\end{eqnarray}
where $\alpha$ is a hyperparameter adjusting the ratio.

\subsection{Updating and Evaluating Logic Rules}
\label{sec:rule_up_eval}
As illustrated above, the collected initial rules are utilized as the seeds to activate our reasoning engine. Nevertheless, they are concept-level and have a limited amount. 
Thus, simply using them to construct the rule base may lack flexibility and generalization. Hence, more qualified logic rules are essential for practical reasoning.

To this end, we propose an \textit{inductive-deductive} policy to maintain and update our logic rule base. During the reasoning, our rule base is not fixed but \textit{scalable} and \textit{renewable}. The logic rule base is first initialized with the human prior logic rules. Additionally, some rule candidates are selected to further enrich our rule base. They mainly have two sources: 1) some of the annotated primitive-activity sets from the HAKE \textit{train set}, and 2) automatically generated rules.

\begin{algorithm}[!ht]
\renewcommand{\algorithmicrequire}{\textbf{Input:}}
\renewcommand{\algorithmicensure}{\textbf{Output:}}
\caption{Evaluating and Updating logic rules}
\label{alg:rule_update}
\begin{algorithmic}[1]
\Require 
Train set T, 
initial model parameter $\Theta^0$, 
maximum rule count $N_r$, 
maximum epoch $K$,
initial rule set $\mathbf{R}^0=\{R_m^0\}_{m=1}^A=\{\{r_i^0\}_{i=1}^{N_m^0}\}_{m=1}^A$,  candidate rule set $\mathbf{R^*}=\{R_m^*\}_{m=1}^A=\{\{r_i^*\}_{i=1}^{N_m^*}\}_{m=1}^A$ ($A$: activity category count, $N_m^0$: initial rule count, $N_m^*$: candidate rule count)
\Ensure
New rule set $\mathbf{R}^{K-1}=\{R_m^{K-1}\}_{m=1}^A$,
trained model parameter $\Theta^K$ 
\For{each $k\in [1,K]$}
    \State Train model on T based on $R_m^{k}$, and update parameter from $\Theta^{k-1}$ to  $\Theta^{k}$  \algorithmiccomment{induction, differentiable}
    \If{$k<K$} \algorithmiccomment{deduction, non-differentiable}
        \For{each $m\in [1,A]$}
        \State $R_m^{k'} = R_m^k\cup R_m^{*}$ \algorithmiccomment{$R_m^{k'}$ is the set of all the candidate rules}
        \State Evaluate the training loss $l_{m_i}$ for each rule $r_{m_i} \in R_m^{k'}, 1 \le i \le n$
        \State Sort $R_m^{k'}$ as $r_{m_{j_1}}, r_{m_{j_2}}, ... r_{m_{j_n}}$, where $l_{m_{j_1}} \le l_{m_{j_2}} \le ... \le l_{m_{j_n}}$
        \State Construct $R_m^{k+1}=\{r_{m_{j_1}}, r_{m_{j_2}}, ..., r_{m_{j_{N_r}}}\}$ 
        \EndFor
    \EndIf
\EndFor
\State \Return{$\mathbf{R}$}
\label{alg-updating}
\end{algorithmic}
\end{algorithm}

First, as the annotated activity-primitive sets of images/videos are \textit{instantiated} rules (\textit{i.e.}, abstract rules and happened cases), some of them may be general and fit other scenarios. Hence, we may find out good rules from the annotations. In practice, we randomly sample the annotated activity-primitive sets as the candidates and convert them into the standard format of the logic rule.

Second, we can also automatically generate rule candidates with our primitive dictionary. As we define about 200 \textit{PaSta} primitives before, the possible rules for one activity can be exaggerated $2^{200}$. Fortunately, each activity has its own primitive bias and just very few possible primitive \textit{compositions} are suitable.
In our implementation, for the $m$-th activity, we first count its co-occurrence with all primitives, \textit{i.e.}, $C_m = \{c_i\}_{i=1}^p$, where $p$ is the number of primitives and $i$ is the primitive index. Then, we impose min-max normalization on $C_m$ and get $C_m^{'} = \{c_i^{'}\}_{i=1}^p$, where $c_i^{'}$ indicates the \textit{occurring probability} of the $i$-th primitive. 
$C_m^{'}$ is the \textit{prior knowledge} for primitives $\rightarrow$ activity relations and can be regarded as a \textit{rule generator}. It can generate rules as $r_m = \{r_{m_i}\}_{i=1}^p$, $r_{m_i} \sim \mathcal{B}(c_i^{'})$. Here, $\mathcal{B}(\cdot)$ is the Bernoulli distribution.
To generate diverse rules, we adopt a different generator based on the prior as
\begin{eqnarray}
    C_{m}^{''}(\beta) = \{c_{i}^{''}(\beta)\}_{i=1}^p
\end{eqnarray}
and we \textit{randomly} make $c_{i}^{''}(\beta)=(1-\beta)*c_i^{'}$ or $c_i^{''}(\beta)=c_i^{'}+\beta*(1-c_i^{'})$ to decrease or increase the primitive probabilities. $\beta \in [0,1]$ is a parameter controlling the ``distance'' between the original and transformed generators, since Kullback-Leibler divergence $\textbf{KL}(\mathcal{B}(c_i^{'}) \parallel \mathcal{B}(c_i^{''}(\beta)))$ increases as $\beta$ increases. In practice, we set $\beta = 0.1 \cdot \theta$ for $\theta \in [0,10]$ and get 11 different generators. We generate 5 rules from each generator and obtain 55 rules for each activity category.

After the rule base is extended with the candidates, we operate the inductive-deductive policy which cycles within each epoch. The procedure is illustrated in Algorithm~\ref{alg:rule_update}. 
Within each epoch, model parameters including the logic operation modules ($NOT(\cdot)$, $OR(\cdot,\cdot)$) are first trained based on the \textbf{fixed rule set}, (induction policy, L2, Algo.~\ref{alg:rule_update}). 
Then, we \textbf{fix model parameters} and evaluated all the candidate rules to \textbf{update the rule set}, \textit{i.e.}, the primitive sets (deduction policy, L3-10, Algo.~\ref{alg:rule_update}). 
Thus, the two parts (logic operation modules and rules) are \textbf{optimized alternately}.
That said, the model is \textbf{updating} when its parameters are trained based on the \textbf{fixed} rule set, while \textbf{frozen} when the rule set is \textbf{evaluating} and \textbf{updating}. In the second stage, though the model is frozen, the loss generated by the model given one candidate rule is used as the basis for the rule selection. 
The two stages are performed alternatively several times in the training phase. From the perspective of the whole training phrase, the logic rules are evaluated and selected with our differentiable reasoning model automatically. One rule selected in the early stage may be dropped in the later stage by the updated model. Meanwhile, a non-selected rule in the early stage may also be selected later by the updated model.
We also illustrate the proposed inductive-deductive policy in Suppl.~Fig.~4.
Through evaluation and updating, we maintain a scalable rule base to better fit the practical scenarios.

\subsection{Unified Activity Inference}
\label{sec:unified_infer}
Though we leverage a different mechanism to infer activities based on primitive representations, our method is not exclusive to previous activity understanding methods~\cite{li2020pastanet,ava,gao2018ican,interactiveness,idn,Lu2016Visual} adopting \textit{instance}-level representations of visual entities (human instances). Thus, as a plug-and-play, we can flexibly integrate our primitive-level representation with the instance-level representation.
We use Eq.~(\ref{eq:res-inst}) to indicate the result from the instance-based method:
\begin{eqnarray}
    \label{eq:res-inst}
    \mathcal{S}_{ins} & = & \mathcal{F}_{ins}(\mathcal{I}, b_h, b_o),
\end{eqnarray}
where $\mathcal{F}_{ins}(\cdot)$ can be any instance-based methods~\cite{ava,gao2018ican,interactiveness,idn}. The loss of instance-level result is denoted as $\mathcal{L}_{ins}$. To obtain the final result, we adopt the late fusion strategy, \textit{i.e.}, compactly fusing the predictions of two levels following $\mathcal{S} = \mathcal{S}_{ins} * \mathcal{S}_{LR}$.
The total loss is constructed as
\begin{eqnarray}
    \label{eq:total-loss}
    \mathcal{L}_{total} & = & \mathcal{L}_{LR} + \mathcal{L}_{ins} + \mathcal{L}_{P},
\end{eqnarray}
where primitive detection can be simultaneously fine-tuned with our unified activity inference.

\section{Experiments}
\label{sec:experiment}

\subsection{Datasets and Metrics}
We conduct experiments on large-scale benchmarks including HICO-DET~\cite{hicodet}, AVA~\cite{ava}, V-COCO~\cite{vcoco}, and Ambiguous-HOI~\cite{djrn}.
HICO-DET~\cite{hicodet} is a benchmark built on HICO~\cite{hico} and provides human-object boxes, which contains 47,776 images and 600 interactions.
AVA~\cite{ava} contains 430 videos with spatio-temporal labels. It includes 80 atomic activities consisting of body motions and HOIs. 
V-COCO~\cite{vcoco} contains 10,346 images with GT instance boxes. 
It contains the 80 object classes from COCO~\cite{coco} and 29 activity categories.
Ambiguous-HOI~\cite{djrn} contains 8,996 images with 25,188 annotated human-object pairs in 87 HOIs (consisting of 48 verbs and 40 objects from HICO-DET~\cite{hicodet}). It is designed to specially examine the ability to process 2D pose and appearance ambiguities.
For HICO-DET~\cite{hicodet}, V-COCO~\cite{vcoco}, and Ambiguous-HOI~\cite{djrn}, we follow the mAP metric: a true positive contains accurate human and object boxes ($IoU>0.5$ regarding GT) and accurate interaction prediction. For a fair comparison, we use 
ResNet-50~\cite{resnet} as the backbone for HAKE and all methods together with their object detectors.
For AVA~\cite{ava}, we follow its frame-mAP metric: a true positive contains an accurate human box ($IoU>0.5$ regarding GT) and accurate activity prediction. Similarly, for fairness, we use object detection from LFB~\cite{lfb} and SlowFast~\cite{SlowFast} for all methods.
Besides, for \textit{PaSta} primitive detection, we adopt the same metric with \cite{hicodet}.

\subsection{Implementation Details}
For enhancing experiments in Sec.~\ref{sec:exp-enhancement}, we use all HAKE image data for HICO-DET and all video data for AVA (the data in the HICO-DET test set and AVA validation set together with the corresponding labels are all \textbf{excluded} to avoid data pollution).
We use \textit{instance-level primitive} labels: each annotated person with the corresponding primitive labels, to train the primitive detector and Activity2Vec and then fine-tune the primitive detector, Activity2Vec, and primitive-based logical reasoning module together on HICO-DET~\cite{hicodet} or AVA~\cite{ava} train set respectively.
When training the reasoning module, the primitive detector is frozen. The pre-training and fine-tuning take 1 M and 2 M iterations respectively. The learning rate is 1e-3 and the ratio of positive and negative samples is 1:4. We use an SGD with momentum (0.9) and cosine decay restart (the first decay step is 80 K). The reasoning module is trained for 5 epochs using Adam with a learning rate of 1e-3. A late fusion strategy is adopted. In Eq.~\ref{eq:reason_loss}, $\alpha=0.2$.

We detail the settings of transfer learning as follows.
On V-COCO~\cite{vcoco} in Sec.~\ref{sec:transfer_experiment}, we {\bf exclude} the images of V-COCO and the corresponding primitive labels in HAKE and use the remaining data (109 K images) for pre-training. 
We use an SGD with 0.9 momenta and cosine decay restarts (the first decay is 80 K). 
The pre-training costs 300 K iterations with the learning rate as 1e-3.
The fine-tuning costs 80 K iterations with a learning rate of 7e-4.
To enhance the transfer effect, besides the primitive logical reasoning result and the instance-level result (from instance-level methods), we also include a \textit{perceptual reasoning result}. That is, with the $f_{P}^V$ extracted with Activity2Vec, we feed it to an MLP and perform the \textit{activity classification}, getting the perceptual reasoning result $\mathcal{S}_{PR}$. Thus, the final result is formulated as $\mathcal{S}=\mathcal{S}_{ins} \times \mathcal{S}_{LR} \times \mathcal{S}_{PR}$.
\begin{table}[!ht]
\centering
\resizebox{0.49\textwidth}{!}{
\setlength\tabcolsep{1pt}{
\begin{tabular}{l|lcc}
\hline
Dataset                  & Method                    & All                   & Rare               \\
\hline
\hline
\multirow{9}{*}{HICO-DET} 
                                & \textbf{GT}-HAKE (GT H-O boxes)     & 62.65 & 71.03 \\
                                & \textbf{GT}-HAKE+rule search (GT H-O boxes)     & \textbf{71.69} & \textbf{82.25} \\
                                        \cline{2-4}
                                        & TIN~\cite{interactiveness}      & 17.03 & 13.42 \\
                                        & TIN~\cite{interactiveness}+HAKE & \textbf{23.22 ($\uparrow$36.3\%)} & \textbf{23.16 ($\uparrow$72.6\%)} \\
                                        & VCL~\cite{vcl}                  & 23.63 & 17.21 \\
                                        & VCL~\cite{vcl}+HAKE             & \textbf{28.30 ($\uparrow$19.8\%)} & \textbf{25.04 ($\uparrow$45.5\%)} \\
                                        & QPIC~\cite{qpic}                & 29.07 & 21.85 \\
                                        & QPIC~\cite{qpic}+HAKE           & \textbf{32.10 ($\uparrow$10.4\%)} & \textbf{27.57 ($\uparrow$26.2\%)} \\
\hline
\multirow{6}{*}{AVA} 
                               & SlowFast~\cite{SlowFast}+\textbf{GT}-HAKE (detection~\cite{SlowFast})  & 42.23 & 30.86 \\ 
                               & \textbf{GT}-HAKE (GT H boxes)  & \textbf{47.27} & \textbf{34.47} \\ 
                               \cline{2-4}
                               & LFB-Res-50-max~\cite{lfb}            & 23.9                   & 5.2  \\
                               & LFB-Res-50-max~\cite{lfb}+HAKE       & \textbf{26.8 ($\uparrow$12.13\%)} & \textbf{7.7 ($\uparrow$48.08\%)}\\
                               & SlowFast~\cite{SlowFast}             & 28.2                   & 9.6    \\
                               & SlowFast~\cite{SlowFast}+HAKE        & \textbf{29.3 ($\uparrow$3.90\%)}  & \textbf{10.4 ($\uparrow$8.33\%)}\\
\hline        
\end{tabular}}}
\vspace{-5px}
\caption{Results (mAP, \%) of enhancing experiments on HICO-DET~\cite{hicodet} and AVA~\cite{ava}. The relative improvements are shown in the brackets.
``GT-HAKE'' means inputting GT primitive to the reasoning module.
``detection'' means using the human (-object) boxes from the detector. ``GT H-O/H boxes'' means inputting the GT human (-object) boxes.
}  
\vspace{-5px}
\label{fig:exp-enhancing}
\end{table}
\begin{table}[!th]
\centering
\resizebox{0.36\textwidth}{!}{
\setlength\tabcolsep{1pt}{
\begin{tabular}{l|lc}
\hline
Dataset                  & Method                    & mAP                   \\
\hline
\hline
\multirow{2}{*}{V-COCO~\cite{vcoco}} & TIN~\cite{interactiveness}      & 54.2                   \\
                                    & TIN~\cite{interactiveness}+HAKE & \textbf{59.7 ($\uparrow$10.1\%)} \\
\hline
\multirow{4}{*}{Ambiguous-HOI~\cite{djrn}} & TIN~\cite{interactiveness}      & 8.22  \\
                                          & TIN~\cite{interactiveness}+HAKE & \textbf{10.56 ($\uparrow$28.5\%)} \\
                                          & DJ-RN~\cite{djrn}               & 10.37 \\
                                          & DJ-RN~\cite{djrn}+HAKE          & \textbf{12.68 ($\uparrow$22.3\%)} \\
\hline
\multirow{2}{*}{AVA~\cite{ava}} & AVA-TF~\cite{ava}       & 11.4 \\
                               & AVA-TF~\cite{ava}+HAKE  & \textbf{15.6 ($\uparrow$36.8\%)} \\
\hline
\end{tabular}}}
\vspace{-5px}
\caption{Transfer learning results on V-COCO~\cite{vcoco}, Ambiguous-HOI~\cite{djrn}, and image-based AVA~\cite{ava} (v2.1).} 
\vspace{-15px}
\label{fig:exp-transfer}
\end{table}
On Ambiguous-HOI~\cite{djrn}, we use the same model and setting used in the enhancing experiment on HICO-DET~\cite{hicodet}.
On image-based AVA~\cite{ava}, as HAKE is built upon still images, we use the \textbf{frames per second} as \textit{still images} for \textbf{image-based} activity detection. We adopt ResNet-50~\cite{resnet} as the backbone and an SGD with a momentum of 0.9. 
The initial learning rate is 1e-2 and the first decay of cosine decay restarts is 350 K.
For a fair comparison, we use the human boxes from LFB~\cite{lfb}.
The pre-training costs 1.1 M iterations and fine-tuning costs 710 K iterations.
Besides, we also adopt an image-based baseline: Faster R-CNN detector~\cite{faster} with ResNet-101~\cite{resnet} provided by the AVA~\cite{ava} website. 
Due to the huge domain gap, to enhance the transfer effect, we used a similar strategy to that for V-COCO~\cite{vcoco}:
besides the logical reasoning result and instance-level result, we include the perceptual reasoning result $\mathcal{S}_{PR}$, getting the final result following $\mathcal{S}=\mathcal{S}_{ins} \times \mathcal{S}_{LR} \times \mathcal{S}_{PR}$. 

\subsection{Results and Comparisons}
\subsubsection{Primitive Detection}
\label{sec:primitive_result}
With extensive primitive annotations, a concise CNN-based primitive detector (ResNet-50) can perform very well.
On HICO-DET\cite{hicodet}, given the same detected human-object boxes~\cite{gao2018ican}, our primitive detector achieves \textbf{30.51} mAP (28.37 (head), 38.18 (arms), 19.48 (hands), 42.06 (hip), 23.84 (legs), 31.12 (feet)) and greatly outperforms the instance-level activity detection performance of state-of-the-art~\cite{vcl} with ResNet-50 (19.43 mAP).
If inputting GT human-object boxes, our primitive detector achieves impressive \textbf{42.22} mAP.
This verifies that primitives can be well learned with the help of our activity-primitive knowledge base.
Similarly, on the image-level HICO\cite{hico}, the performance gap between primitive and activity recognition is also large. HAKE achieves \textbf{55} mAP on primitive recognition and an activity recognition state-of-the-art~\cite{Fang2018Pairwise} achieves 40 mAP.

\begin{figure}[H]
	\begin{center}
		\includegraphics[width=0.35\textwidth]{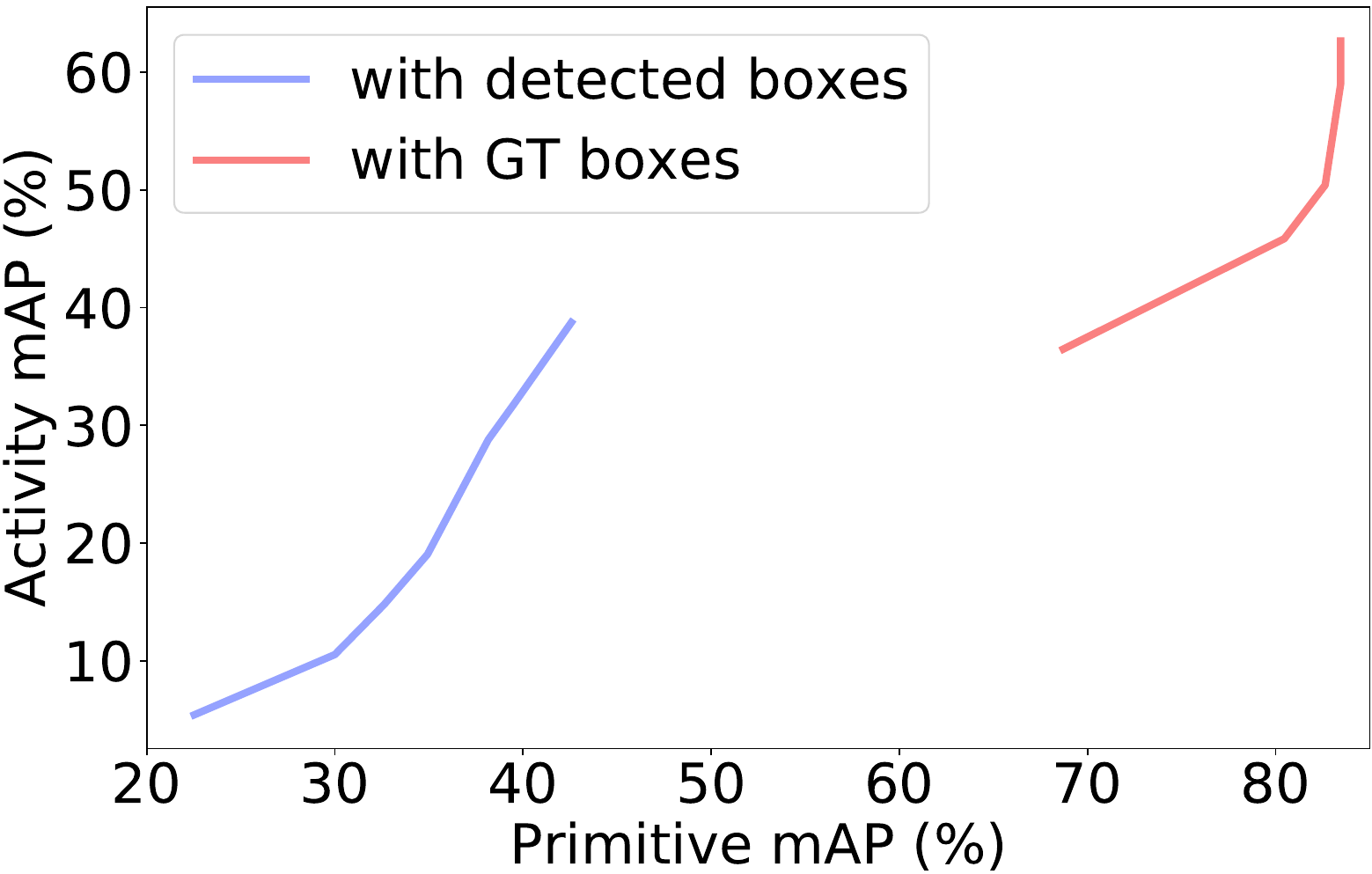}
	\end{center}
	\vspace{-15px}  
	\caption{Positive correlation between the primitive detection quality and activity detection performance (w/o unified inference) on HICO-DET~\cite{hicodet}. Detected boxes mean the detection from iCAN~\cite{gao2018ican}.}
	\label{figure:noise_det}
\vspace{-10px}
\end{figure}

\subsubsection{HAKE-Enhanced Activity Detection}
\label{sec:exp-enhancement}
We conduct the challenging instance-level activity detection on HICO-DET~\cite{hicodet} and AVA~\cite{ava} which needs to locate active humans/objects and classify activities simultaneously, to show the enhancing effect of HAKE.
State-of-the-arts are chosen to compare and cooperate with HAKE. 

\noindent{\bf HICO-DET.}
Results are shown in Tab.~\ref{fig:exp-enhancing}. 
On HICO-DET~\cite{hicodet}, HAKE significantly boosts previous instance-level methods especially on the Rare sets, bringing \textbf{9.74} mAP improvement upon TIN~\cite{interactiveness} and strongly proving the efficacy of learned primitive information. 
Notice that even upon the methods with much stronger backbone and object detection like VCL~\cite{vcl} and QPIC~\cite{qpic}, the relative improvements are still considerable (both over 10\%).
Moreover, GT-HAKE (the upper bound of HAKE with \textbf{perfect} primitive detection) outperforms the state-of-the-art significantly, indicating the \textit{potential} of our logical reasoning.
Besides, on the image-level activity recognition benchmark HICO~\cite{hico}), we achieve higher than \textbf{50} mAP on 206 image-level HOIs (\textbf{206/600}). 

\noindent{\bf AVA.}
Meanwhile, on the AVA~\cite{ava} validation set, HAKE also brings significant improvements over previous methods. It boosts the detection performance of a considerable number of activities, especially the 20 most \textbf{rare} activities.

More detailed results are provided in Suppl.~Tab.~3,~4.

\subsubsection{Evaluating the Transfer Ability of HAKE}
\label{sec:transfer_experiment}
In three challenging transfer learning experiments, HAKE also shows efficacy.
Results are shown in Tab.~\ref{fig:exp-transfer}.

\noindent{\bf V-COCO.}
We select state-of-the-art as baselines and adopt the metric $AP_{role}$~\cite{vcoco} (requires accurate human-object boxes and activity prediction). 
With domain gap, HAKE still brings \textbf{5.5} mAP (\textbf{10.1\%}) improvement upon TIN~\cite{interactiveness}.

\noindent{\bf Ambiguous-HOI.}
On more difficult Ambiguous-HOI, HAKE also improves the performance by \textbf{2.34} and \textbf{2.31} mAP (\textbf{28.5\%} and \textbf{22.3\%} relative improvements) upon~\cite{interactiveness,djrn}.

\noindent{\bf Image-based AVA.} 
Both image- and video-based methods cooperating with HAKE achieve impressive improvements, even when our model is trained \textbf{without temporal information}. 
Considering the \textit{huge domain gap} (movies in AVA~\cite{ava}) and \textit{unseen activities}, these results strongly prove the generalization ability of HAKE again.

More detailed results are provided in Suppl.~Tab.~6-8. 
We also visualize some activity cases to exhibit the generalization of HAKE in Suppl.~Fig.~5.

\subsubsection{Analyzing the Upper Bound of HAKE}
Next, we reveal the potential of HAKE: 

\begin{figure}[H]
	\begin{center}
		\includegraphics[width=0.35\textwidth]{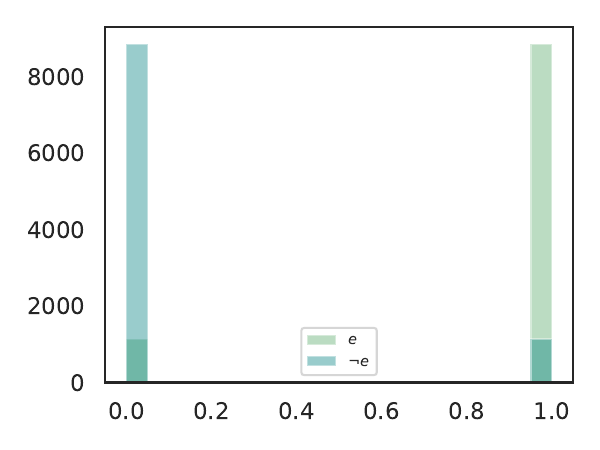}
	\end{center}
	\vspace{-15px}
	\caption{Verification of the logic operations. Distribution of the $True$/$False$ in sampled embeddings, which shows the well-distinguishable ability of HAKE for two logical states.}
	\label{figure:eval_reasoning}
\vspace{-10px}
\end{figure}
\begin{table}[H]
\begin{center}
\resizebox{0.45\textwidth}{!}{
\begin{tabular}{cccc}
\hline  
Expression & Accuracy & Expression & Accuracy \\
\hline  
$x \neq \neg x$ & 0.9411 & $x=\neg \neg x$ & 0.9426 \\
$x \vee x = x $ & 0.8854 & $x \vee \neg x = T$ & 0.9425 \\
$T \vee T = T $ & 0.9870 & $T \vee F = T$   & 0.9960 \\
$F \vee T = T $ & 0.9990 & $F \vee F = F$   & 0.7950 \\
$T \vee T \vee T = T $ & 0.9870 & $T \vee T \vee F = T$   & 0.9940 \\
$T \vee F \vee F = T $ & 0.9790 & $T \vee F \vee F = F$   & 0.7180 \\
\hline
\end{tabular}}
\end{center}
\vspace{-10px}
\caption{Accuracy of different logic expressions. $x$ stands for random embedding, and $T, F$ stand for $True, False$ respectively.}
\label{tab:eval_reasoning}
\vspace{-5px}
\end{table}

1)	\textit{Ground Truth (GT) Primitive}. We input GT primitives with random noise to the reasoning engine, \textit{i.e.}, less noise indicates better primitive detection. 
Given the ratio $mr \in [0,1]$ and total $n$ pairs in the test set, we randomly choose $n*mr$ pairs and replace the GT primitives with evenly distributed noise. In practice, we set $mr=0, 0.005, 0.01, 0.05, 0.1, 0.2, 0.5$ and calculate primitive/activity mAP.  
As shown in Fig.~\ref{figure:noise_det}, less noise (better primitive detection) results in better activity detection.
In Tab.~\ref{fig:exp-enhancing}, on the challenging HICO-DET~\cite{hicodet}, the upper bounds are \textbf{45.52} (+QPIC~\cite{qpic}, detection~\cite{qpic}) and \textbf{62.65} (GT human-object boxes) mAP, which are significantly superior to the state-of-the-arts (about \textbf{29} mAP~\cite{qpic} and \textbf{44} mAP~\cite{idn}).
Here, detection~\cite{qpic} indicates using the detected human-object boxes from~\cite{qpic}.
On AVA~\cite{ava}, the upper bound of HAKE is also impressive, \textit{i.e.}, 42.23 (+SlowFast~\cite{SlowFast}, detection~\cite{SlowFast}) and 47.27 (GT human boxes) mAP, which also largely outperform the SlowFast~\cite{SlowFast} (about 28 and 34 mAP).

2)	\textit{GT Primitives + Million Rules}. To verify the completeness of primitive-semantic space bridging, we input GT primitives and search for the rules that suit the current data distribution best (Suppl.~Sec.~4.2).
This operation further boosts performance. After searching 1 M rules for each possible activity of each sample, 
HAKE achieves impressive \textbf{71.69} (GT H-O boxes) mAP on HICO-DET~\cite{hicodet}, which vastly outperforms the above results and verifies the firm guarantee and potential of our logical reasoning.

\subsubsection{Evaluating the effectiveness of logical modules}
To verify the logical reasoning ability of our logical modules, we examine them with the logic expressions on the test set. 

\begin{figure}[!ht]
\begin{center}
    \includegraphics[width=0.5\textwidth]{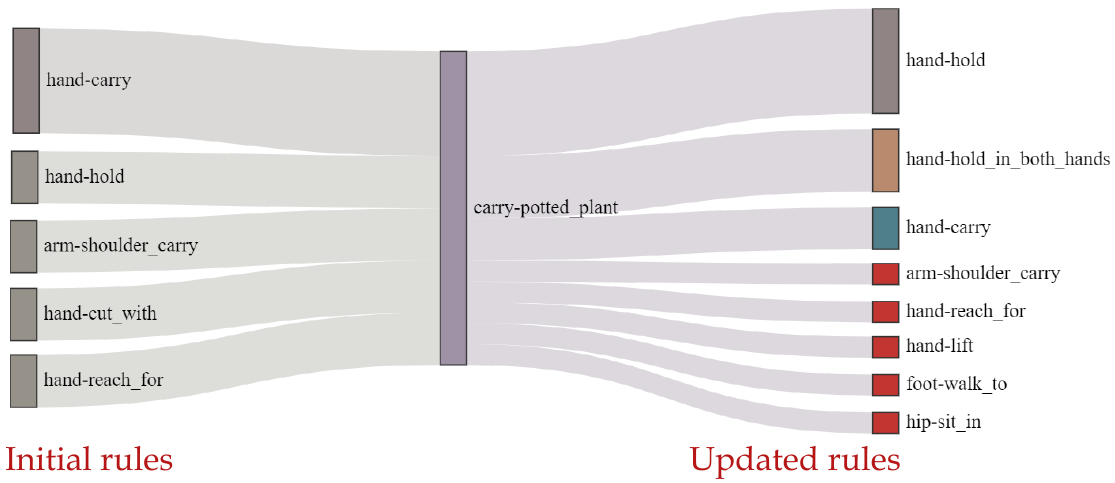}
\end{center}
    \vspace{-15px}
	\caption{The correlation between \textit{PaSta} and logic rules before and after the updating for activities \textit{hug horse} and \textit{carry potted\_plant}.}
	\label{Fig:rule_c}
	\vspace{-10px}
\end{figure}
\begin{figure}[!ht]
\begin{center}
    \includegraphics[width=0.5\textwidth]{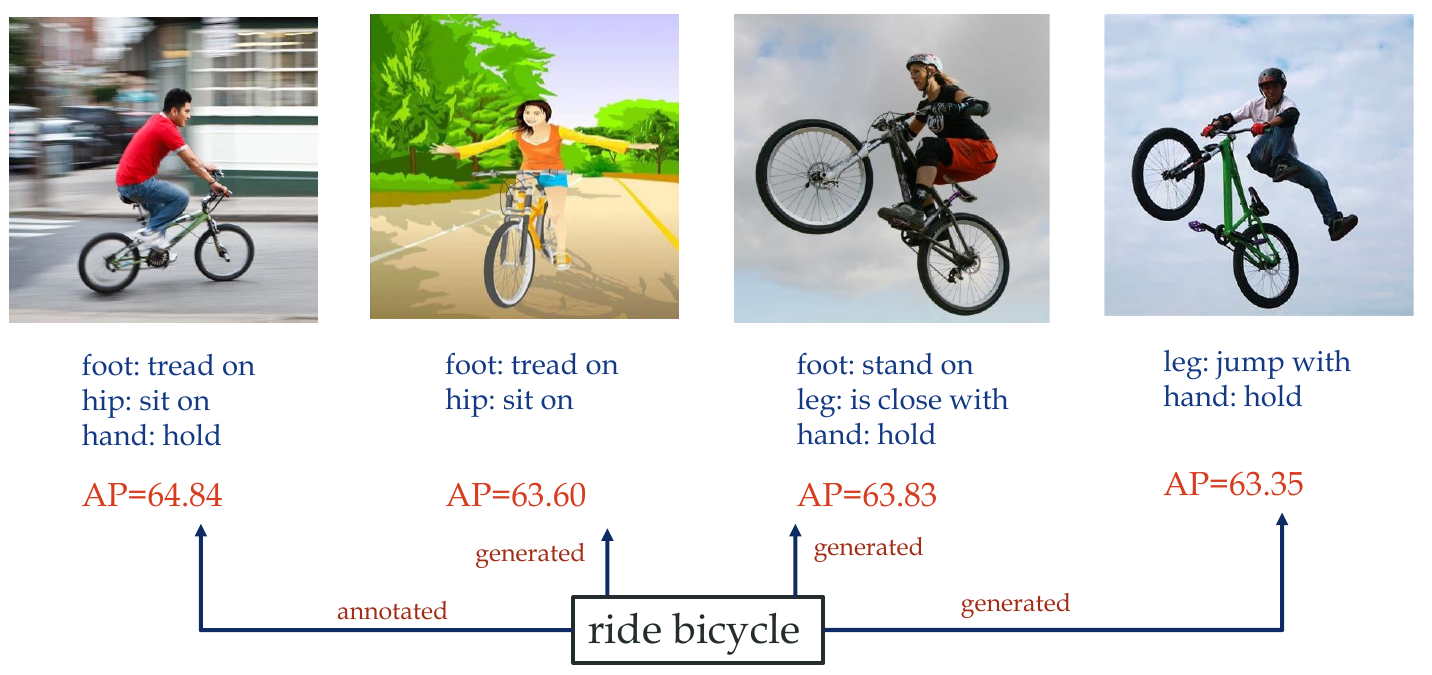}
\end{center}
    \vspace{-15px}
	\caption{Annotated rules and generated rules for \textit{ride bicycle} class and their performances (AP) on the HICO-DET~\cite{hicodet} test set.}
	\label{Fig:rule_f}
	\vspace{-15px}
\end{figure}

Specifically, we randomly sample 20 K event embeddings in the test set from $e_{P}$, $e_{A}$, and $e_{m}$ as the inputs of logic expressions.
We first feed them into the binary discriminator $\mathcal{J}(\cdot)$ to obtain their $True/False$ probabilities. 
As shown in Fig.~\ref{figure:eval_reasoning}, the sampled event embeddings show obvious logical characteristics, as most embeddings have logical probabilities close to either 0 or 1, which means they represent $True/False$ without ambiguity. If the embeddings are sampled randomly from the \textit{whole} vector space rather than the test set event space, the probabilities distribution will be Gaussian instead of bipolar.

To further demonstrate the effectiveness of logical modules, we choose a threshold $t_l\geq 0.5$ as the criterion to select embeddings to represent $True/False$. We select embeddings $e$ such that $\mathcal{J}(e)>t_l$ as $True$ and $\mathcal{J}(e)<t_l$ as $False$ then feed them into the logical modules to evaluate the ability of logical reasoning. It should be noted the threshold $t_l$ actually sets a \textit{stricter} standard since any probabilities higher than $0.5$ should be regarded as $True$. The accuracy of every logic expression is shown in Tab.~\ref{tab:eval_reasoning}.
It is evident that in the given \textit{specific event space}, our logic modules have the ability of logical reasoning since the accuracy of all binary and ternary logic expressions exceeded 50\% significantly. 
It should also be noted that the logical modules are not ideal and restricted in our defined task space. The modules tend to judge the embedding as $True$ and lose the ability to infer with multiple-layer logic expressions. This is expected since the backpropagation-driven DNN modules tend to take a shortcut and the inputs of logical modules are not balanced. 
Despite this, the results of experiments still demonstrate the good logical reasoning ability of our logical reasoning engine in the \textit{activity event space}.

\subsubsection{Analyzing Logic Rules}
Fig.~\ref{Fig:rule_c} shows the correlation between \textit{PaSta} and rules before and after the updating for \textit{hug horse} and \textit{carry potted\_plant}. Two endpoints are more related if they are connected with a wider stream. After updating, HAKE has learned more diverse and suitable \textit{PaSta} related to the activity.

For most activities, the farther away from the generated rule distribution from the \textit{human prior} distribution, the worse the performance. However, the results of some activities are just the opposite, \textit{e.g.}, \textit{ride bicycle} (green line).
This is in detail demonstrated in Fig.~\ref{Fig:rule_f}. The four pictures present various situations for \textit{ride bicycle}. 
The 1st image from the left shows a common case. However, people may also ride a bicycle without hands holding the handles (the 2nd), without the hip sitting on the seat (the 3rd), or with feet jumping from the bicycle (the 4th).
Such various situations are very difficult and costly to fully annotate case by case. However, our rule generator can \textit{discover} these corner cases by evaluating the generated logic rules. As shown in Fig.~\ref{Fig:rule_f}, besides the annotated 1st rule, the other three rules are all generated and adopted by our inductive-deductive reasoning policy.
We further measure their performances on HICO-DET~\cite{hicodet} and find that these ``corner'' rules also fit the data well, achieving 63.60, 63.83, and 63.35 AP respectively compared with the annotated one (64.84 AP).

Due to the pape limit, we provide more evaluations and analyses in Suppl.~Sec.~4, \textit{e.g.}, the evaluations of logic rules, logic reasoning, completeness of primitive space, \textit{etc}.

\subsection{Ablation Study}
We design ablation studies on Ambiguous-HOI~\cite{djrn} under transfer learning and GT-HAKE (GT H-O boxes) settings to avoid the influence of primitive detection. For more ablation studies, please refer to Suppl.~Sec.~6.3.

\noindent{\bf Language Feature.}
To verify the ability of the linguistic feature, we replace the primitive Bert feature in Activity2Vec with Gaussian noise, Word2Vec, and GloVe on GT-HAKE mode (64.21 mAP, w/o rule update). The results are all worse (60.55, 62.17, 62.43 mAP).
The worst performance of Gaussian noise shows the efficacy of the linguistic features.

\noindent{\bf Logic Rules Complexity.}
We change the upper limit of the adopted logic rule number to test its influence on the final performance. 
In the beginning, the inference ability of HAKE improves as human prior knowledge extends. However, as the rule count further increases, the performance falls. The possible reason is that the rule base is ``overfitted'' and more rules would hurt the generalization ability of HAKE. For more details, please refer to Suppl.~Sec.~4.1.

\noindent{\bf Rule Updating Policy.} 
1) Without the evaluation and updating of logic rules, the performance degrades from 68.20 mAP to 64.21 mAP, verifying the efficacy of the inductive-deductive policy.
2) When evaluating the rules, we can select better rules according to either lower train loss or higher performance (mAP) on the train set data. We find that they perform basically the same (68.20 and 68.15 mAP respectively).
3) During updating, the candidate rules have two sources: the annotated primitive-activity pairs from the HAKE train set and automatically generated rules. With annotated or generated \textit{only}, the performance falls to 66.89 and 66.96 mAP respectively. 

\subsection{SCR Test}
\begin{figure}[!ht]
\begin{center}
\vspace{-10px}
    \includegraphics[width=0.35\textwidth]{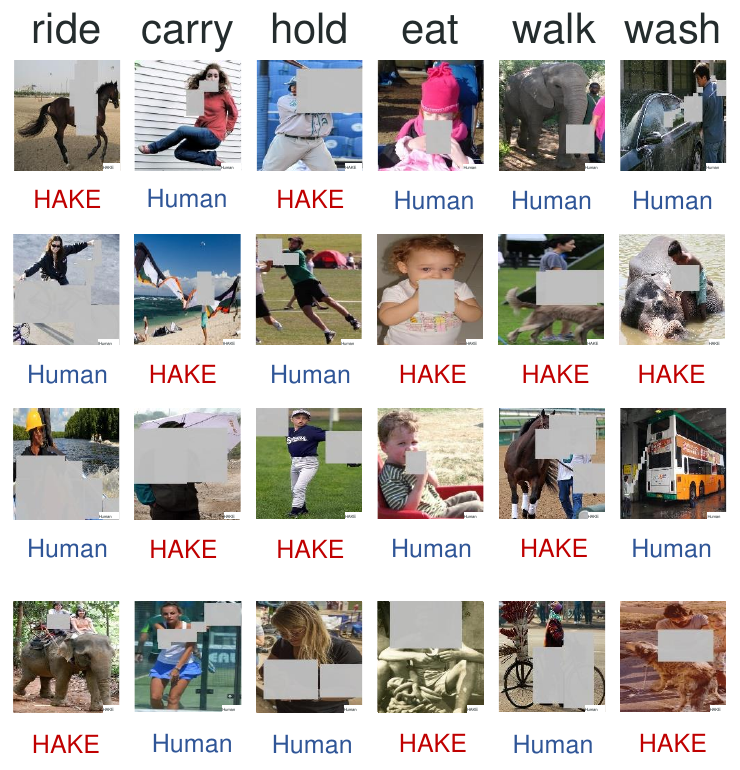}
\end{center}
    \vspace{-12px}
	\caption{Masking results from HAKE and humans in the SCR Test. The verbs are given at the top. The source of masking is marked below the sub-image. Two sets of maskings are very similar and difficult to distinguish even by human participants (59.55\% accuracy).}
	\label{Fig:SCR}
	\vspace{-15px}
\end{figure}
Human intelligence can create things with few samples, but evaluating activity understanding systems via creation is difficult. 
For another, SCR measures the area ratio of key semantic regions that humans can quickly discover. Similarly, HAKE can detect primitives, so we can utilize it to discover key semantics. Hence, to verify that HAKE has gained the knowledge to recognize the logic relations between primitives and activities, we propose an SCR test based on the SCR principle, \textit{i.e.}, \textbf{understanding-via-elimination} which compares the \textit{semantics elimination effect} of humans and HAKE. That said, given an image, humans and HAKE would do their best to mask the \textit{minimum} and \textit{equivalent} pixels to eliminate the action semantics, \textit{i.e.}, making other human participants cannot recognize the activities. 

We conduct a test on 1,000 images of 105 activities, human participants, and HAKE attempt to mask the \textit{smallest} and \textit{equivalent} box regions to eliminate key semantics.
Finally, we calculate the recognition performance of other participants on the unmasked and two masked image sets from humans and HAKE. 
For the unmasked images, the human recognition performance is 90+\%.
Besides, for the two masked sets, the performances are close: \textbf{35.6}\% (human masking) and \textbf{39.5}\% (HAKE masking), indicating that HAKE can eliminate key semantics well with a similar image masking ratio to humans.
Some examples are visualized in Fig.~\ref{Fig:SCR}. The texts under the images are the sources of the maskings. We can find out that HAKE achieves good \textit{key regions localization} and \textit{recognition abilities}. 

Moreover, an additional test to determine whether HAKE’s masking is \textit{indistinguishable} from humans’. Through a user study, the accuracy of masking source discrimination (humans or HAKE) is only \textbf{59.55}\%.
Thus, HAKE can create masking very similar to humans', thus can robustly transform the activity information of image space to primitive space. 
For more details, please refer to Suppl.~Sec.~7.

\section{Conclusion}
HAKE explores a new insight to advance activity understanding. We worked with 702 participants to release a large-scale and fine-grained knowledge base as fuel for reasoning. HAKE is, to the best of our knowledge, the \textit{first} and \textit{largest} publicly available activity dataset with \textit{conceptual} and \textit{logical descriptions}, which yields a great performance boost, particularly for few-shot learning. Furthermore, our primitive dictionary and logic rule base are scalable and can be easily adapted to novel scenarios. We will open the website upload port for users to upload their primitives and rules to continually enrich HAKE.

From the SCR test, HAKE shows its significant ability to locate key semantics via primitive detection and reasoning, ensuring that primitive space can faithfully embed the visual activity information.
And the reasoning engine bridges primitive semantic space well. 
All of these factors give us the hope that HAKE can approach the success of object recognition. To prove this, we analyzed its upper bound on challenging tasks, where it shows significant superiority.
With the enrichment of primitive dictionaries and rule base, we believe HAKE will lay the foundation for an interpretable and applicable system for areas related to human survival and development, \textit{e.g.}, ambient intelligence~\cite{feifeinature}. 
For example, detecting patient activities in the ICU to monitor accidents such as falls, planning diet, and treatment according to the amount of exercise, and monitoring sleep quality by detecting turnover, getting up, \textit{etc}.

HAKE also harnesses concept learning with \textit{compositional generalization}. 
First, as the compositions of primitives, activities usually differ \textit{locally}, just as molecules are composed of atoms and differ in the type, number, and compound mode of atoms. We can manipulate primitives and transform \textit{global} activities into novel categories. This elegant characteristic will advance zero-shot learning. 
Second, inductive reasoning extracting rules from raw data and deductive reasoning verifying rules are practically promising for combining deep learning and symbolic reasoning. Considering the practicability of activity understanding, we hope HAKE will be a good platform based on real-world data for cognition analysis and causal inference~\cite{pearl2016causal}.

\section{Acknowledgment}
This work was supported by the National Key Research and Development Project of China (No. 2021ZD0110700), Shanghai Municipal Science and Technology Major Project (2021SHZDZX0102), Shanghai Qi Zhi Institute, and SHEITC (2018-RGZN-02046).
\ifCLASSOPTIONcaptionsoff
  \newpage
\fi

\bibliographystyle{IEEEtran}
\bibliography{egbib}

\newpage

\begin{appendices}
\section{Details of the Bottleneck Analysis of Direct Mapping} 
To investigate the difference between object patterns and activity patterns, we invite participants and divide them into two groups: the first group would mask the key areas of object or activity images and make efforts to make other participants cannot recognize the objects or activities again. Meanwhile, the second group would be given the masked images from the first group and recognize them. The masked images that cannot be recognized would be noted as \textbf{successful cases}. 
The object and activity image sets have the same number of images. In detail, we collect 742 images for object recognition covering 58 objects from COCO~\cite{coco} and 745 images for activity recognition covering 92 activities from HICO-DET~\cite{hicodet} to ensure diversity. To avoid biases, we balance the images of object and activity classes: about 13 images for each object category and 10 images for each activity category.

\begin{figure*}[ht]
 	\begin{center}
 		\includegraphics[width=\textwidth]{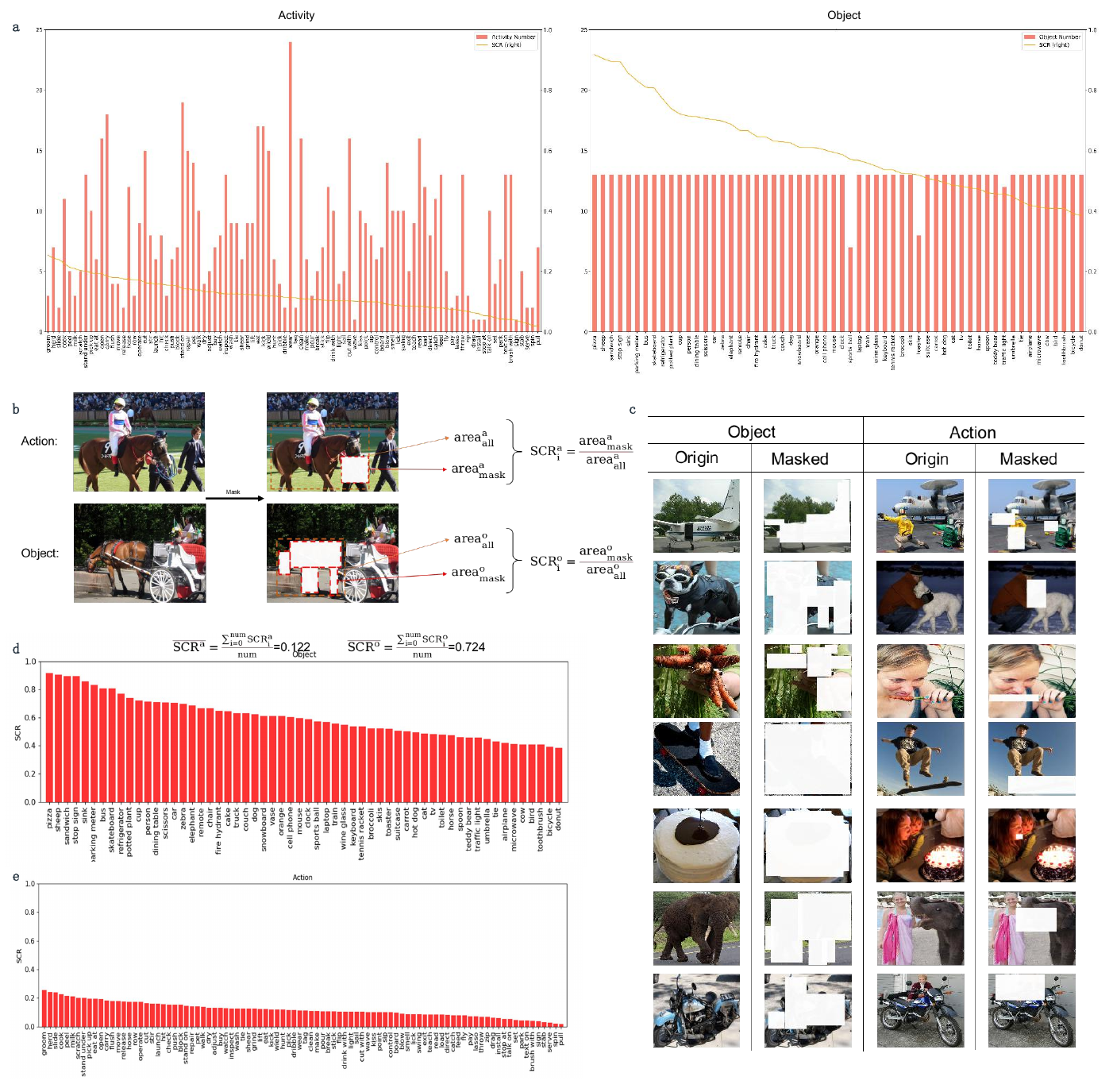}
 	\end{center}
  \caption{Details of the bottleneck analysis based on SCR. 
  \textbf{a.} Image numbers of different activity/object categories in an SCR descending order. 
  \textbf{b.} The definition and illustration of SCR measurement. 
  \textbf{c.} Some examples of SCR test results on object/activity images.
  \textbf{d.} Specific SCRs of different object categories.
  \textbf{e.} Specific SCRs of different activity categories.}
    \label{fig:SCR Test}
\end{figure*}

As shown in Suppl.~Fig.~\ref{fig:SCR Test}a, the image number of activity classes varies more dramatically than object classes. 
This is due to more severe long-tailed distribution~\cite{hicodet} on activity classes.
Comparatively, the long-tailed distribution of object classes is relatively mild because the dataset~\cite{coco} we adopt has been manually modulated to balance the object classes. 
All masks are added within the tight bounding boxes containing all the activity/object entities, \textit{i.e.}, whole human body (human-object pair), or object. Moreover, all the lengths and widths of these boxes are greater than 200 pixels to provide clear visual information.
After the masking by humans, SCR is calculated as the area ratio between the mask and tight box in \textbf{successful cases} (Suppl.~Fig.~\ref{fig:SCR Test}b). 
From the results shown in Suppl.~Fig.~\ref{fig:SCR Test}c, we can find that the activity images have much smaller SCRs than the object images, which strongly verifies the huge difference between activity and object patterns. Specifically, the activity images have the average SCR of \textbf{0.122}, while the object images have the average SCR of \textbf{0.610}.
To make the results more vivid, we visualized the SCRs of each activity/object class in Suppl.~Fig.~\ref{fig:SCR Test}d,e. 

It can be easily discovered that the activity classes have much lower SCRs (0.0-0.3). And the SCRs of object classes are between 0.4-0.9. 
As a result, accurately capturing the key activity semantic regions is much more difficult than object recognition. Direct mapping thus requires more training data to be able to locate the concentrated activity semantics and thus hinders the performance. 
Hence, we need different paradigms for these two different tasks. 
To this end, we propose HAKE for activity understanding.

\section{Knowledge Base Constructing Details}
This section details the construction processes of the HAKE knowledge base and the logic rule base.

\subsection{Primitive Collection} 
\label{sec:pasta-collection}
We first introduce the collection of activity primitives. HAKE seeks to explore the common knowledge of activity primitives as atomic elements to infer activities. 

{\bf Primitive Definition.} 
For part states (\textit{PaSta}), we decompose the human body into \textit{ten} parts, namely \textit{head, two arms, two hands, hip, two legs, two feet}. Then, \textit{PaSta} will be assigned to describe these body parts. For example, the primitive of \textit{hand} can be \textit{hold-something (sth)} or \textit{push-sth}, the primitive of \textit{head} can be \textit{watch-sth}, \textit{eat-sth}. 
After exhaustively reviewing the collected 122 K+ images and 74.7 hours (299) of videos, we find that the descriptions of any human parts can be concluded into limited classes (about 100). Notably, a person may have more than one activity simultaneously. Thus each part can have multiple primitives too. Furthermore, body parts can also convey pose information such as \textit{hand-wave} and \textit{waist-bend}.
As for objects, we focus on the most common 80 objects from the COCO\cite{coco} dataset.
As for scenes, we adopt the 400 scenes from~\cite{zhou2017places}.

{\bf Data Collection.}
For generalization, we collect human-centered activity images by crowd-sourcing (with rough activity label) as well as existing well-designed datasets~\cite{hico,hicodet,vcoco,openimages,hcvrd,pic} which are structured around a rich semantic ontology, diversity, and variability of activities. All their annotated persons and objects are extracted for our construction. 
For AVA~\cite{ava}, we collect the active persons' annotations from its available 299 15-minute videos (train and validation sets of AVA 2.1).
Finally, we collect more than 350 K+ images/frames of diverse activity categories. 

{\bf Activity Labeling.} 
Activity categories of HAKE are chosen according to the most common \textit{daily human activities, interactions with object/person}. Referred to the hierarchical structure\cite{activitynet} of activity, common activities in existing datasets~\cite{hico,vcoco,hcvrd,openimages,ava,activitynet,MPII,pic} and crowd-sourcing labels, we select 156 activities including \textbf{human-object interactions} (also including human-human interactions) and \textbf{body only motions}. We detail the 156 activities in Suppl.~Tab.~\ref{tab:action}.
We first clean and reorganize the annotated persons and objects from existing datasets and crowd-sourcing. Then, we annotated the bounding boxes of active persons and interacted COCO 80 objects in the rest images. As for AVA~\cite{ava} videos, since its diverse and moving objects, we follow previous works~\cite{ava,SlowFast,lfb} and focus on the active persons only.

\begin{table*}[htb]
  \centering
  \begin{tabular}{p{60pt}p{410pt}}
    \hline    
    HOIs & adjust, assemble, block, blow, board, break, brush with, board gaming, buy, carry, catch, chase, check, chop, clean, clink glass, close, control, cook, cut, cut with, dig, direct, drag, dribble, drink with, drive, dry, eat, eat at, enter, exit, extract, feed, fill, flip, flush, fly, fight, fishing, give sth to sb, grab, greet, grind, groom, hand shake, herd, hit, hold, hop on, hose, hug, hunt, inspect, install, jump, kick, kiss, lasso, launch, lick, lie on, lift, light, listen to sth, listen to a person, load, lose, make, milk, move, open, operate, pack, paint, park, pay, peel, pet, play musical instrument, play with sb, play with pets, pick, pick up, point, pour, press, pull, push, put down, put on, race, read, release, repair, ride, row, run, sail, scratch, serve, set, shear, shoot, shovel, sign, sing to sb, sip, sit at, sit on, slide, smell, smoke, spin, squeeze, stab, stand on, stand under, stick, stir, stop at, straddle, swing, tag, take a photo, take sth from sb, talk on, talk to, teach, text on, throw, tie, toast, touch, train, turn, type on, walk, wash, watch, wave, wear, wield, work on laptop, write, zip\\
    \hline
    Body Motions & bow, clap, climb, crawl, dance, fall, get up, kneel, physical exercise, swim\\
    \hline
  \end{tabular}
  \caption{Activities in HAKE. HOIs indicate the interactions between a person and an object/person.}
  \label{tab:action}
\end{table*}

\begin{table*}[htb]
  \centering
  \begin{tabular}{p{60pt}p{410pt}}
    \hline    
    Head    & eat, inspect, talk with, talk to, close with, kiss, raise up, lick, blow, drink with, smell, wear, listen to, no activity\\
    \hline
    Arm  & carry, close to, hug, swing, crawl, dance, material art, no activity  \\
    \hline
    Hand & hold, carry, reach for, touch, put on, twist, wear, throw, throw out, write on, point with, point to, use sth to point to, press, squeeze, scratch, pinch, gesture to, push, pull, pull with, wash, wash with, hold in both hands, lift, raise, feed, cut with, catch with, pour into, crawl, dance, martial art, no activity \\
    \hline
    Hip & sit on, sit in, sit beside, close with, bend, no activity \\
    \hline
    Thigh & walk with, walk to, run with, run to, jump with, close with, straddle, jump down, walk away, bend, kneel, crawl, dance, material art, no activity\\
    \hline
    Foot & stand on, step on, walk with, walk to, run with, run to, dribble, kick, jump down, jump with, walk away, crawl, dance, fall down, martial art, no activity \\ 
    \hline
    \hline
    Objects & airplane, apple, backpack, banana, baseball bat, baseball glove, bear, bed, bench, bicycle, bird, boat, book, bottle, bowl, broccoli, bus, cake, car, carrot, cat, cell phone, chair, clock, couch, cow, cup, dining table, dog, donut, elephant, fire hydrant, fork, frisbee, giraffe, hair drier, handbag, horse, hot dog, keyboard, kite, knife, laptop, microwave, motorcycle, mouse, orange, oven, parking meter, person, pizza, potted plant, refrigerator, remote, sandwich, scissors, sheep, sink, skateboard, skis, snowboard, spoon, sports ball, stop sign, suitcase, surfboard, teddy bear, tennis racket, tie, toaster, toilet, toothbrush, traffic light, train, truck, tv, umbrella, vase, wine glass, zebra\\
    \hline
  \end{tabular}
  \caption{Activity primitives consist of human body part states (\textit{PaSta}) and objects.}
  \label{tab:pasta}
\end{table*}

{\bf Body Part Box for \textit{PaSta}.} 
To locate the human parts, we use pose estimation~\cite{fang2017rmpe} to obtain the joints of all annotated persons. Then we follow Fang~\textit{et.~al.}~\cite{Fang2018Pairwise} and generate \textit{ten} body part boxes. Estimation errors are addressed manually to ensure high-quality annotation. Each part box is centered with a joint, and the box size is pre-defined by scaling the distance between the joints of the neck and pelvis. A joint with confidence higher than 0.7 will be seen as visible. When not all joints can be detected, we use \textit{body knowledge-based rules}. That said, if the neck or pelvis is invisible, we configure the part boxes according to other visible joint groups (head, main body, arms, legs), \textit{e.g.}, if only the upper body is visible, we set the size of the hand box to twice the pupil distance.

\begin{figure*}[ht]
 	\begin{center}
 		\includegraphics[width=\textwidth]{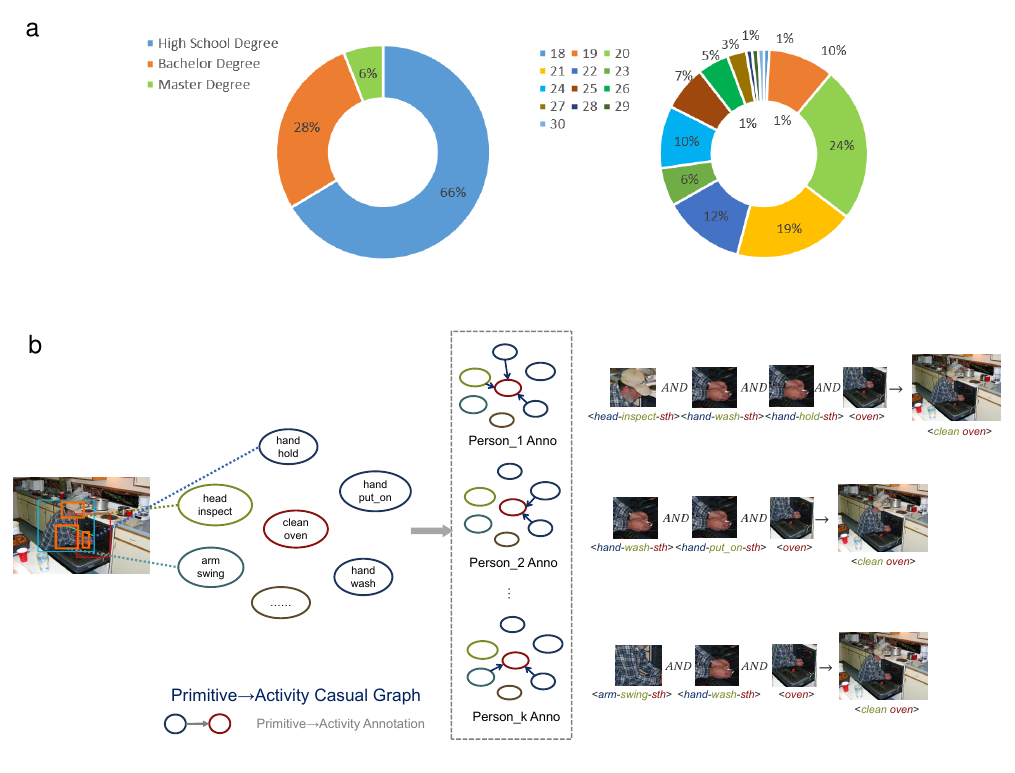}
 	\end{center}
 	\caption{HAKE construction details. 
 	\textbf{a.} Participant statistics of HAKE construction and experiments. Basic participant background information including education and age. 
 	\textbf{b.} The primitive-activity initial rule annotation. Activity-Primitive annotation to generate the causal graph (logic rules), \textit{i.e.}, which primitives occurring simultaneously would cause an activity.}
 	\label{figure:casual_anno}
\end{figure*}

{\bf Primitive Annotation.} 
We carry out the annotation by crowd-sourcing. 
As for object primitives, we directly annotate the interacted objects in human-object interactions with their 80 classes\cite{coco} and tight boxes.
As for scenes, here, we use a model pre-trained on Place365~\cite{zhou2017places} to classify the scenes in images/frames and manually fix the wrong predictions. We would not always use scene information in reasoning since the user study concludes that scenes usually cannot decide the activities and may in turn import noise/bias in inference. For example, the spurious correlation between \textit{play football} and \textit{grassland} would bias the reasoning.
For the more complex human body \textit{PaSta}, the process is as follows:
1) First, we choose the \textit{PaSta} categories considering the \textit{generalization}. For example, about 200 human body part states (\textit{PaSta}~\cite{lu2018beyond,li2020pastanet} from WordNet~\cite{miller1995wordnet} are chosen to form a primitive dictionary, \textit{e.g.}, \textit{foot-kick-sth}. If a part does not have any active states, we depict it as \textit{no\_activity}).
2) Second, to find the most common \textit{PaSta} that can work as the \textit{transferable activity knowledge}, we invite 200 annotators from different backgrounds to annotate 10 K images and 50 videos of 156 activities with our primitive dictionary. For example, given an activity \textit{ride bicycle}, they may describe it as \textit{hip-sit\_on-sth}, \textit{hand-hold-sth}, \textit{foot-step\_on-sth}, \textit{etc}. When the participants cannot explain their decisions with the current dictionary, they should propose new \textit{PaSta}. If a proposed \textit{PaSta} is frequently used by other participants, it will be formally included.
3) Based on their annotations, we use the Normalized Point-wise Mutual Information (NPMI)~\cite{church1990word} to calculate the co-occurrence between activities and \textit{PaSta} in the dictionary. Finally, \textbf{93} \textit{PaSta} with the highest NPMI values are chosen.
4) Using the annotations of 10 K images and 50 videos as seeds, we automatically generate the initial \textit{PaSta} labels for all of the rest data. Thus the other 408 annotators are asked to revise the annotations according to the specific human states in images.
5) Considering that a person may have multiple activities, for \textit{each} activity, we annotate its corresponding ten \textit{PaSta} respectively. Then we combine all sets of \textit{PaSta} from all activities. Thus, a part can also have multiple states, \textit{e.g.}, in activity \textit{eating while talking}, the head has \textit{PaSta} \textit{head-eat-sth}, \textit{head-talk\_to-sth} and \textit{head-look\_at-sth} simultaneously.
6) To ensure quality, each image or video clip will be annotated twice and checked by automatic procedures and supervisors. We cluster all labels and discard the outliers to obtain robust agreements.
We detail the \textit{Pasta} classes in Suppl.~Tab.~\ref{tab:pasta}.

Finally, HAKE includes 
\textbf{122 K+} images (\textbf{247 K+} persons, \textbf{345 K} instance activities, \textbf{220 K+} object primitives, and \textbf{7.4 M+} \textit{PaSta} primitives) 
and \textbf{299} 15-min videos (\textbf{74.7} hours; one second per frame, \textbf{234 K+} frames in total; \textbf{426 K+} persons, \textbf{1.0 M+} instance activities, and \textbf{18.8 M+} \textit{PaSta} primitives). 

With this large-scale activity-primitive knowledge base, we find that primitives can be well \textbf{learned}. 
A shallow and concise CNN trained with a part of HAKE data can easily achieve about \textbf{55} mAP in \textit{image-level primitive recognition} on HICO~\cite{hico} and \textbf{30} mAP in \textit{instance-level primitive detection} on HICO-DET~\cite{hicodet}.
Meanwhile, deeper and sophisticated state-of-the-art~\cite{Fang2018Pairwise,vcl} can only achieve about 40 and 20 mAP on \textit{image-level activity recognition} and \textit{instance-level activity detection}.
Moreover, primitives can be well \textbf{transferred}. To verify this, we conduct transfer learning experiments on Ambiguous-HOI~\cite{djrn}, V-COCO~\cite{vcoco}, and image-based AVA~\cite{ava}, \textit{i.e.}, first train a backbone model to learn the knowledge of HAKE and then use it to infer the activities of unseen data, even unseen activities. Results show that primitives can be well transferred and boost the performance greatly (main test Tab.~2).

\subsection{Annotation Analysis}
There were \textbf{702} participants involved in the construction of HAKE. They have various backgrounds. Thus we can ensure annotation diversity and reduce biases. The basic background information is shown in Suppl.~Fig.~\ref{figure:casual_anno}a.
We also list the selected 156 activities which cover both person-object/person interactions, body-only motions, and the selected primitives in our dictionary (93 \textit{PaSta} and 80 objects) in Suppl.~Sec.~\ref{sec:action}.

\subsection{Initial Logic Rule Collection}
\label{sec:rule-collection}
To initialize our logic rule base, we invite 5 participants to annotate the human prior logic rules according to their understanding of human activities. Given the activity \textbf{concepts} only (activity class names), they are asked to describe these concepts with our primitive concepts from the dictionary. Hence, the collected logic rules would be the abstract \textit{concept-level} and learned from the participants' daily life experiences without bias from images/videos.
As illustrated in Suppl.~Fig.~\ref{figure:casual_anno}b, participants are asked to annotate the \textit{primitives $\rightarrow$ activity} causal graphs, \textit{i.e.}, which primitives existing simultaneously can cause the effect (specific activities).
To deduce one activity, the causes (primitives) are annotated as \textit{one}, and the primitive nodes in the graph would have arrows pointing to this activity node. On the contrary, the unrelated primitive nodes are annotated as \textit{zero}.
For example, for activity \textit{clean oven}, the possible cause-effect results are \textit{hand-wash-sth} $\land$ \textit{oven}$\rightarrow$\textit{clean oven}, or \textit{hand-put\_on-sth$\land$ hand-wash-sth} $\land$ \textit{oven}$\rightarrow$\textit{clean oven}.
After the annotation, for the $m$-th activity, we obtain its initial logic rule set $R_{m}=\{r_{i}\}_{i=1}^{l_m}$, where $r_{i} \in \{0,1\}^{p}$, $r_{i}$ is a binary sequence, $p$ indicates the number of primitives, and $l_m$ is the rule count.
This initial logic rule base covers most human daily activities and thus carries the convincing human prior knowledge of activities. It would act as a good starting or seed for the subsequent automatic logic rule discovery and neuro-symbolic reasoning.

{\bf Logic Rule Aggregate.}
After annotation, we get $m_0$ rules for each activity.
For the $i$-th activity,  its rule set $R_{i}=\{r_{i}^{j}\}_{j=1}^{m_0}$, $r_{i}^{j} \in \{0,1\}^{q}$, where $r_{i}^{j}$ is the $j$-th rule for the $i$-th activity, $q$ is the number of primitives.
Based on the annotation, we aggregate the logic rules and summarize the activity expertise base.

First, we exclude the repetitive logic rules $r_{i}^{j}$ in $R_{i}$ and get $R_{i}^{'}=\{r_{i}^{j}\}_{j=1}^{m}$.
Next, we calculate $r_{i}^{mean}=\sum_{j=1}^{m} r_{i}^{j}$ and rank $\{r_{i}^{j}\}_{j=1}^{m}$ according to the Euclidean/Cosine Distance between $r_{i}^{mean}$ and $r_{i}^{j}$. Then, we sample $n$ rules with equal intervals in the rank list.
Finally, for the $i$-th activity, we aggregate $n$ logic rules as its expertise base. The sampling strategy increases the diversity of cognition and provides a better generalization. On the basis of prior logic rule collection, we can construct a convincing and professional expertise base.

\section{Method Details}
\subsection{Primitive-Based Logical Reasoning}
\begin{figure}[!ht]
\begin{center}
    \includegraphics[width=0.48\textwidth]{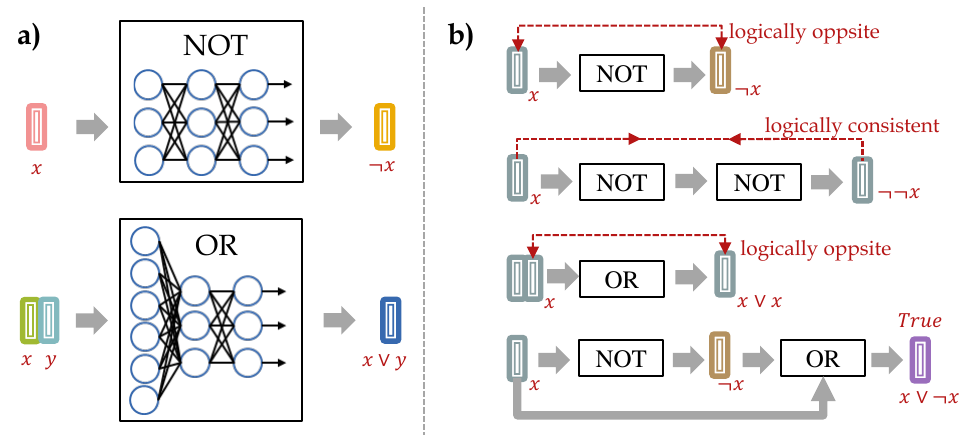}
\end{center}
	\caption{Illustration of the logic operation modules.
	a) The logic operation modules $NOT(\cdot)$, $OR(\cdot; \cdot)$ are implemented as MLPs that are reusable for all vectors in the event space; 
	b) The logic operation modules are constrained by the basic laws of logic.}
	\label{fig:logic_module}
\end{figure}

The logic operation modules $NOT(\cdot)$, $OR(\cdot; \cdot)$ are implemented as Multi-Layer Perceptrons (MLPs) that are reusable for all vectors in the event space (Sec.~4.1.2 of the main text). To better clarify the details, we further illustrate them in Suppl.~Fig.~\ref{fig:logic_module}a.
In terms of the $NOT(\cdot)$ module, the size of the input and output are the same.
In terms of the $OR(\cdot; \cdot)$, the size of the input tensor is double the output size.

Besides, Suppl.~Fig.~\ref{fig:logic_module}b shows how these modules are constrained by the basic laws of logic: 1) NOT negation: $\neg$ TRUE = FALSE, 2) NOT double negation: $\neg \neg x = x$, 3) OR idempotence: $x\vee x = x$, 4) OR complementation: $x\vee (\neg x) =$ TRUE, where $x \in \mathcal{X}$ refers to the event vector, and $\mathcal{X}$ is the event vector space. This explains $\mathcal{L}_{reg}$ in Sec.~4.3 of the main text in detail.

\subsection{Updating and Evaluating Logic Rules}
\begin{figure}[!ht]
\begin{center}
    \includegraphics[width=0.48\textwidth]{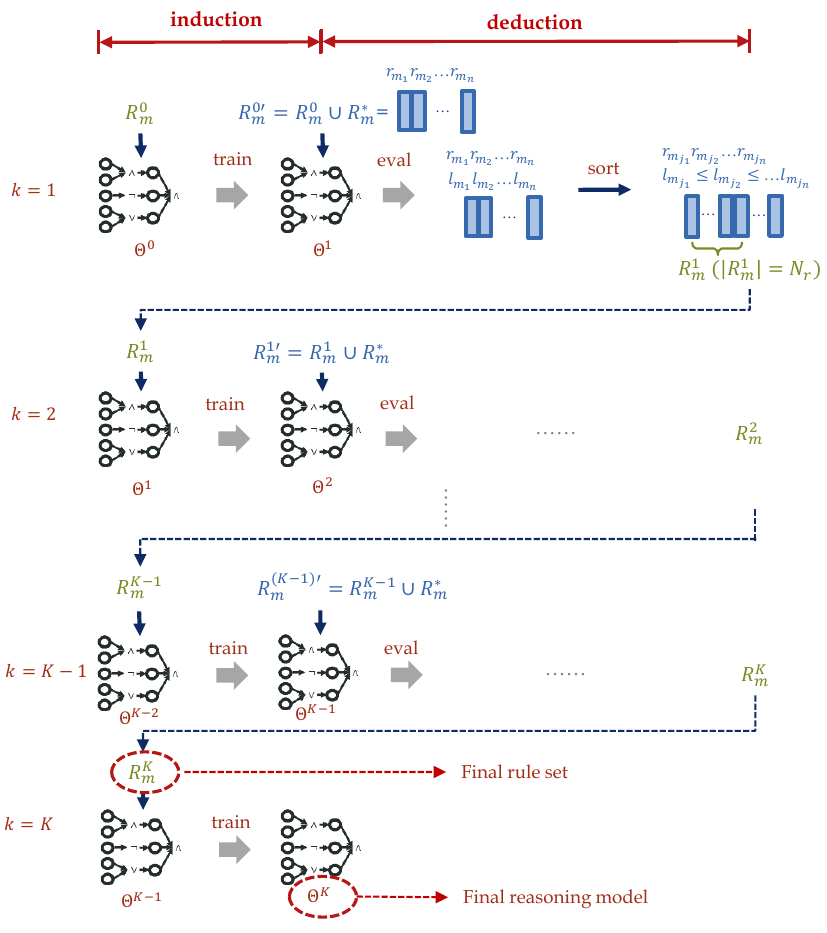}
\end{center}
	\caption{Illustration of the evaluating and updating of logic rules.}
	\label{fig:rule_update}
\end{figure}

Here, we further give a figure to illustrate the proposed inductive-deductive policy in Suppl.~Fig.~\ref{fig:rule_update}.

\subsection{Synthesizing Logic Rules}
In Sec.~4.3 of the main text, we mentioned that there are two fusion strategies for deriving the final prediction, late fusion and early fusion. The framework uses these two fusion strategies in different stages.
Here, we provide some explanations for such a usage choice.

We first compare late/early fusion strategies.
In the late fusion strategy, we obtain the explicit \textbf{predictions} from each rule and then average the predictions for the final prediction. In the early fusion strategy, our model directly combines the \textbf{embeddings} generated from multiple rules without explicit predictions.
Therefore, the late fusion is more interpretable and easier to scale to more rules.

In rule discovery and updating, our model evaluates rules from their explicit predictions, and it processes a relatively large number of rules. Thus, we adopt the late fusion.
After that, we freeze the updated rule base and finetune the whole reasoning module via early fusion. This is mainly based on experiment results--early fusion improves the model performance with more flexibility to accept rule errors. Without finetuning via early fusion, the performance degrades from 68.20 mAP to 66.85 mAP on Ambiguous-HOI under transfer learning and GT-HAKE (GT H-O boxes) setting.

\section{More Analyses}
\label{sec:suppl_analyses}
In this section, we provide more analyses and details about HAKE and the experiments.

\subsection{Analysis of Logic Rules}
\label{sec:rule_analysis}
\subsubsection{Rule Complexity}
For each activity category, we first exclude the repetitive rules and count the collected ones. 
Interestingly, different activities usually have different \textbf{rule complexities} in annotation: activities with fewer ambiguities such as \textit{hold baseball\_bat} usually have fewer different initial rules, but the rules of ambiguous activities such as \textit{pet cat} differ greatly, \textit{e.g.}, \textit{arm-hug-sth} $\land$ \textit{head-kiss-sth} $\land$ \textit{cat}$\rightarrow$\textit{pet cat}, \textit{hand-scratch-sth} $\land$ \textit{hand-hold-sth} $\land$ \textit{head-inspect-sth} $\land$ \textit{cat} $\rightarrow$\textit{pet cat}.
Activity \textit{hold baseball\_bat} finally gets 13 rules out of 945 samples (ratio=1.4\%) while  \textit{pet cat} gets 27 rules out of 345 samples (ratio=7.8\%). This follows our common sense that the latter activity is more complex than the former one.

\subsubsection{Activity logical complexity}
In addition to the collected rules, we also analyze the \textit{learned} rules from our reasoning engine. 
As illustrated in Suppl.~Fig.~\ref{fig:analysis}a, when trained with a different number of rules, HAKE achieves different performances. 
The performances on HICO-DET~\cite{hicodet} Full Set and Rare Set both reach the plateaus when the rule count is about $10$-$15$, which proves the validity of our primitive-based logical reasoning. HAKE only needs \textbf{a small quantity} of human prior knowledge to achieve state-of-the-art, while previous methods require hundreds of instance annotations per activity category. 
If we continue to increase the rule count, the performances of some activity categories may still be boosted but most of them start to fall, the reason might be that their rule complexities are ``overfitted'' and more rules are redundancy and even in turn obstruct the reasoning.
We also explore the relationships between performance and rule count for several activity categories in Suppl.~Fig.~\ref{fig:analysis}b, which further verifies the ``saturation'' timings are different for different activities.

\subsubsection{Evaluating and Updating}
After evaluating and updating, we finally got 4,090 logic rules for 156 activity categories.
On HICO-DET~\cite{hicodet}, we get 3,746 rules for all activity categories. Among them, 80.8\% of rules come from the activity-primitive annotation, and 19.2\% of rules come from the generated candidates.
Suppl.~Fig.~\ref{fig:analysis}d shows the initial and updated logic rules for activity \textit{kiss cat} and \textit{direct bus}. After updating, more reasonable rules are learned and absorbed by our rule base (orange text), while the ``worse'' rules which cannot perform qualified are filtered out (purple text, \textit{e.g.}, \textit{head-inspect-sth} for activity \textit{direct bus}). 
These strongly verify the effectiveness of our inductive-deductive strategy.

\subsubsection{Generated Rules.}
\begin{figure*}[ht]
	\begin{center}
		\includegraphics[width=0.95\textwidth]{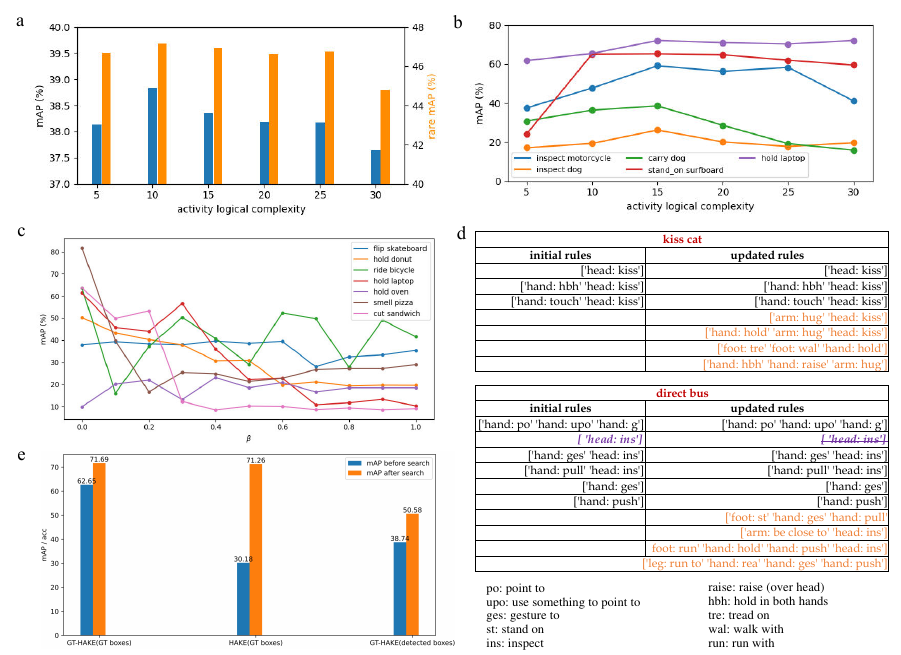}
	\end{center}
	\caption{Analysis of Logic Rules. 
	\textbf{a.} Performance on Full Set and Rare Set of HICO-DET~\cite{hicodet} with changing activity logical complexity. Blue indicates the Full set performance and orange indicates the Rare set performance. 
	\textbf{b.} Performances of different activity categories on HICO-DET~\cite{hicodet} with changing activity logical complexity. 
	\textbf{c.} Performance of different activities on HICO-DET~\cite{hicodet} with changing $\beta$. 
	\textbf{d.} The initial and updated logic rules for activities \textit{kiss cat} and \textit{direct bus}. 
	\textbf{e.} Results of rule search. Performance on HICO-DET~\cite{hicodet} \textbf{after/before (a/b)} giving the larger rule search range. ``GT-HAKE'' indicates inputting GT primitives to the reasoning engine. 
    Relatively, ``HAKE'' indicates that we use the detected primitives from our primitive detector in reasoning.
    ``GT boxes'' means inputting ground truth human-object boxes. 
    Meanwhile, ``detected boxes'' means using the detected boxes from the object detector~\cite{faster,gao2018ican} in reasoning.} 
	\label{fig:analysis}
\end{figure*}

When increasing the $\beta$ in the rule generator, the KL divergence between the human prior distribution and the distribution of generated rules would be larger. We also analyze to study how this affects performance. Suppl.~Fig.~\ref{fig:analysis}c illustrates the performance on HICO-DET~\cite{hicodet} with $\beta$ changing ($\beta \in [0,1]$).  
For most activities, the farther away from the generated rule distribution from the human prior distribution, the worse the performance. However, there are a few cases that indeed work, \textit{e.g.}, \textit{ride bicycle} (green line).

\subsection{Completeness of Primitive Space}
\label{sec:suppl_eva_completeness}
An important advantage of reasoning activity from primitives is the \textbf{limited} set of primitives can powerfully describe complex and \textbf{numerous} activities. 
Then a natural question follows: how to evaluate the completeness of the primitive set? Given suitable rules, how accurately can an activity be inferred based on the existing primitives? 
To answer these questions, we search and optimize logic rules to estimate the \textit{upper bound} of the neuro-symbolic reasoning.

\begin{figure}[!th]
	\begin{center}
		\includegraphics[width=0.4\textwidth]{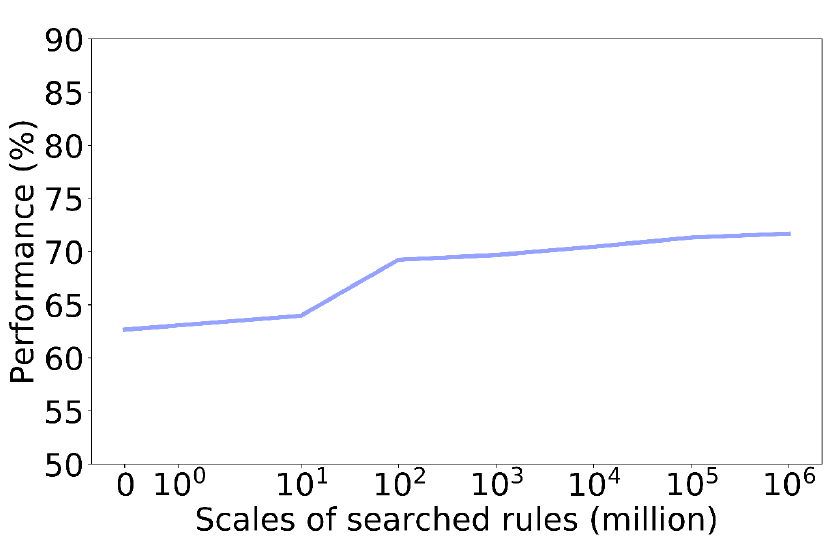}
	\end{center}
	\vspace{-0.5cm}
	\caption{GT primitives and extensive rule search result in better performance (mAP, \%) on HICO-DET~\cite{hicodet}.} 
	\label{fig:search_rule_scale}
\end{figure}

Given $q$ primitives, one logic rule $r_k \in \mathbf{R}_q$, $\mathbf{R}_q=\{0,1\}^q$ totally has $2^q$ kinds of possible permutations. 
The model would achieve its upper bound when the \textbf{most suitable} rules are selected for each activity inference of each human-object pair.  
As demonstrated before, during training, the model evaluates and updates the logic rules to select more qualified rules for practical reasoning. 
However, this is limited, because all human-object pairs \textit{share} the same rules for reasoning. Besides, the rules are selected from the annotated primitive-activity sets or are automatically generated based on some principles, which are quantitatively limited compared to the huge solution space $\mathbf{R}_q=\{0,1\}^q$. 
Thus, the potential of our model is not fully exploited and this is a \textbf{trade-off} question between performance and computation complexity. Under the setting of $q=76$ (only using the HOI-related \textit{PaSta} primitives), $\mathbf{R}_q=\{0,1\}^q$ is too computationally expensive to exhaust all possible rules.
Specifically, approximately \textit{$2 \times 10^{14}$ years} are needed to exhaust all the $2^{76}$ logic rules for \textit{each pair} in HICO-DET.
For AVA, as it contains much more video frames, the process would be unaffordable.

In practice, given the $i$-th human-object pair and the $m$-th activity, each time a rule $r_k$ is randomly selected, we record its corresponding output prediction $p_{imk} \in [0,1]$ for the $m$-th activity based on the pre-trained model. The process is incremental. Since enumerating all the $2^{76}$ rules is computationally unreachable, we make an effort to search and sample about \textbf{1 M} rules for each possible activity of each human-object pair (all testing pairs in the HICO-DET test set). 
Then, we select the ``best rule'' from all 1 M rules based on the recorded predictions. The distance $(p_{imk}-l_{im})^2$ between the prediction $p_{imk}$ and the binary label $l_{im}=0/1$ is calculated and the rule $r_{k^*_{im}}$ with $k^*= \mathop{argmin}_{k} (p_{imk}-l_{im})^2$ is selected as the final rule.
Finally, we use $p_{imk^*_{im}}$ as the final prediction. 

Suppl.~Fig.~\ref{fig:analysis}e show the results with the above rule search policy. 
With more searched rules, our model gains an increase in performance under all settings. 
Specifically, with GT primitives and human-object boxes both, rule search makes HAKE achieve \textbf{71.69} mAP on HICO-DET\cite{hicodet}.
This is much higher than the same upper bound setting w/o rule search (\textbf{62.65} mAP).
Interestingly, when conducting a large-scale rule search and selecting the ``best rules'', the performance gap between using the GT or detected primitives would be much smaller (71.69 and 71.26 mAP). The possible reason may be that even the detected primitives are not always accurate, rule search can also find enough good rules given a part of detected existing primitives and boost the performance. But that is the result of an expensive rule search. In practice, future work would be practical to focus on the improvement of primitive detection and reasoning implementation to tap the potential of HAKE.
Moreover, as illustrated in Suppl.~Fig.~\ref{fig:search_rule_scale}, the performance on HICO-DET\cite{hicodet} improves with increasing searched rules. Meanwhile, the growth also gradually slows down. Thus, the results in Suppl.~Fig.~\ref{fig:analysis}e based on 1 M rules are a sound upper bound estimation of our neuro-symbolic reasoning.
All of these strongly prove the validity of inferring activities via primitives and the potential of HAKE. Once suitable rules are selected, the HAKE reasoning engine can accurately reason out the activities given defined primitives.

Here, ``best rules'' refer to searching for optimal rules to adapt a specific activity understanding dataset. We use it as an ideal metric to estimate the upper bound of our model and verify the completeness of the primitive set.
When activities are expressed via primitives, there are usually many possible compositions of primitives for a complex activity concept. Thus, we develop rule evaluation and updating to improve flexibility and generalization. 
The GT Primitives + Million Rules part is an extension: given the \textbf{rules most suitable to current data distribution} via exhaustive search, how well will our model perform?
The experiment shows HAKE achieves an impressive 71.69 (GT H-O boxes) mAP on HICO-DET, which vastly outperforms the original 62.65 mAP. 
Despite under an ideal setting, it shows our model's \textit{upper bound} with best-fitted rules, thus verifying that an activity can be well inferred based on the defined primitives.

\section{Activities and Primitives in HAKE}
In this section, we detail the classes of activities and primitives in HAKE.

{\bf Activities in HAKE}
\label{sec:action}
HAKE includes 156 different everyday human activities, including human-object/human interactions (HOIs) and body-only motions.
The included categories are: 
adjust, assemble, block, blow, board, break, brush with, board gaming, buy, carry,catch, chase, check, chop, clean, clink glass, close, control, cook, cut, cut with, dig,direct, drag, dribble, drink with, drive, dry, eat, eat at, enter, exit, extract, feed, fill, flip, flush, fly, fight, fishing, give something to somebody, grab, greet, grind, groom, hand shake, herd, hit ,hold, hop on, hose, hug, hunt, inspect, install, jump, kick, kiss, lasso, launch, lick,lie on, lift, light, listen to something, listen to a person, load, lose, make, milk, move, open, operate, pack, paint, park, pay, peel, pet, play musical instrument, play with somebody, play with pets, pick, pick up, point, pour, press, pull, push, put down, put on, race, read, release, repair, ride, row, run, sail, scratch, serve, set, shear, shoot, shovel, sign, sing to somebody, sip, sit at, sit on, slide, smell, smoke, spin, squeeze, stab, stand on, stand under, stick, stir, stop at, straddle, swing, tag, take a photo, take something from something, talk on, talk to,teach, text on, throw, tie, toast, touch, train, turn, type on, walk, wash, watch, wave, wear, wield, work on laptop, write, zip, bow, clap, climb, crawl, dance, fall, get up, kneel, physical exercise, swim.

{\bf Primitives in HAKE}
\label{sec:pasta}
HAKE primitives contain 93 human body part states (\textit{PaSta}) and 80 common objects.

\noindent(1) 14 primitives are defined for the head. They are: eat, inspect, talk with, talk to, close with, kiss, raise up, lick, blow, drink with, smell, wear, listen to, no activity.

\noindent(2) 8 primitives are defined for the arm. They are: carry, close to, hug, swing, crawl, dance, material art, no activity.

\noindent(3) 34 primitives are defined for hand. They are: hold, carry, reach for, touch, put on, twist, wear, throw, throw out, write on, point with, point to, use sth to point to, press, squeeze, scratch, pinch, gesture to, push, pull, pull with, wash, wash with, hold in both hands, lift, raise, feed, cut with, catch with, pour into, crawl, dance, martial art, no activity.

\noindent(4) 6 primitives are defined for the head. They are: sit on, sit in, sit beside, close with, bend, no activity.

\noindent(5) 15 primitives are defined for the leg. They are: walk with, walk to, run with, run to, jump with, close with, straddle, jump down, walk away, bend, kneel, crawl, dance, material art, no activity.

\noindent(6) 16 primitives are defined for the foot. They are: stand on, step on, walk with, walk to, run with, run to, dribble, kick, jump down, jump with, walk away, crawl, dance, fall down, martial arts, no activity.

\noindent(7) 80 objects are included in HAKE, which are: airplane, apple, backpack, banana, baseball bat, baseball glove, bear, bed, bench, bicycle, bird, boat, book, bottle, bowl, broccoli, bus, cake, car, carrot, cat, cell phone, chair, clock, couch, cow, cup, dining table, dog, donut, elephant, fire hydrant, fork, frisbee, giraffe, hair drier, handbag, horse, hot dog, keyboard, kite, knife, laptop, microwave, motorcycle, mouse, orange, oven, parking meter, person, pizza, potted plant, refrigerator, remote, sandwich, scissors, sheep, sink, skateboard, skis, snowboard, spoon, sports ball, stop sign, suitcase, surfboard, teddy bear, tennis racket, tie, toaster, toilet, toothbrush, traffic light, train, truck, tv, umbrella, vase, wine glass, zebra.

\section{Detailed Results and Analysis of Experiments}
\subsection{HAKE-based Enhancing}
We conduct the challenging instance-based activity detection experiments on HICO-DET~\cite{hicodet} and AVA~\cite{ava} which need to locate the active humans/objects and classify the activities simultaneously, to show the enhancing effect of HAKE.
More detailed results are reorganized and shown in Suppl.~Tab.~\ref{tab:hico-det} and Suppl.~Tab.~\ref{tab:ava-enhance}. 

\begin{table}[!t]
\centering
\resizebox{0.5\textwidth}{!}{
\begin{tabular}{l  c  c  c  c  c  c}
\hline
         & \multicolumn{3}{c}{Default}  &\multicolumn{3}{c}{Known Object} \\
Method         & Full & Rare & Non-Rare  & Full & Rare & Non-Rare \\
\hline
\hline
InteractNet~\cite{Gkioxari2017Detecting} & 9.94  & 7.16  & 10.77 & - & - & -\\
GPNN~\cite{qi2018learning}  & 13.11 & 9.34  & 14.23 & - & - & -\\
iCAN~\cite{gao2018ican}     & 14.84 & 10.45 & 16.15 & 16.26 & 11.33 & 17.73 \\
TIN~\cite{interactiveness}  & 17.03 & 13.42 & 18.11 & 19.17 & 15.51 & 20.26 \\
VCL~\cite{vcl}              & 23.63 & 17.21 & 25.55 & 25.98 & 19.12 & 28.03 \\
QPIC~\cite{qpic}            & 29.07 & 21.85 & 31.23 & 31.68 & 24.14 & 33.93 \\
\hline
\hline
HAKE                            & 19.52 & 17.29 & 20.19 & 21.99 & 20.47 & 22.45 \\
TIN~\cite{interactiveness}+HAKE & 23.22 ($\uparrow$36.3\%) & 23.16 ($\uparrow$72.6\%) & 23.24 & 26.16 & 26.54 & 26.04\\ 
VCL~\cite{vcl}+HAKE             & 28.30 ($\uparrow$19.8\%) & 25.04 ($\uparrow$45.5\%) & 29.28 & 30.99 & 27.41 & 32.05\\
QPIC~\cite{qpic}+HAKE           & \textbf{32.10} ($\uparrow$10.4\%) & \textbf{27.57} ($\uparrow$26.2\%) & \textbf{33.46} & \textbf{34.40} & \textbf{29.55} & \textbf{35.85} \\
\hline
GT-HAKE (detection~\cite{gao2018ican})           & 38.74 & 47.11 & 36.24 & 39.52 & 47.44 & 37.16 \\
GT-HAKE (detection~\cite{qpic})                  & 41.63 & 47.04 & 40.02 & 43.40 & 48.79 & 41.79 \\
QPIC~\cite{qpic}+GT-HAKE (detection~\cite{qpic}) & \textbf{45.52} & \textbf{47.60} & \textbf{44.90} & \textbf{46.98} & \textbf{49.00} & \textbf{46.37} \\
GT-HAKE (GT boxes)  & 62.65 & 71.03 & 60.15 & - & - & -\\
GT-HAKE+rule search (GT boxes) & \textbf{71.69} & \textbf{82.25} & \textbf{68.54} & - & - & - \\
\hline
\end{tabular}}
\caption{Results of enhancing experiment on HICO-DET~\cite{hicodet}.} 
\label{tab:hico-det}
\end{table}

\begin{table}[!t]
\begin{center}
\resizebox{0.45\textwidth}{!}{
\begin{tabular}{ccc}
\hline  
Method & Full & Rare-20  \\
\hline  
LFB-Res-50-max~\cite{lfb}    & 23.9 & 5.2\\
LFB-Res-101-nl-3l~\cite{lfb} & 26.9 & 7.8\\
SlowFast~\cite{SlowFast}     & 28.2 & 9.6\\
\hline
LFB-Res-50-max~\cite{lfb}-HAKE        & 26.8 ($\uparrow$12.13\%) & 7.7 ($\uparrow$48.08\%) \\
LFB-Res-101-nl-3l~\cite{lfb}-HAKE     & 27.9 ($\uparrow$3.72\%)  & 8.6 ($\uparrow$10.26\%) \\ 
SlowFast-Res-101~\cite{SlowFast}-HAKE & \textbf{29.3} ($\uparrow$3.90\%)  & \textbf{10.4} ($\uparrow$8.33\%) \\ 
\hline
GT-HAKE (detection~\cite{SlowFast}) & 39.55 & 25.73\\
SlowFast~\cite{SlowFast}+GT-HAKE (detection~\cite{SlowFast}) & 42.23 & 30.86\\
GT-HAKE (GT boxes) & \textbf{47.27} & \textbf{34.47}\\
\hline
\end{tabular}}
\end{center}
\caption{Results of enhancing experiment on AVA~\cite{ava}. As the rule search on AVA is too expensive considering the enormous frames of AVA, we did not conduct the +search rule (GT primitive) test.} 
\label{tab:ava-enhance}
\end{table}

As shown in the main text, the performance gain is considerably lower on QPIC than on the other two methods.
There are two major reasons.
\textbf{First}, The three methods adopted \textbf{different backbones}. 
TIN and VCL used ResNet-50, while QPIC adopted the transformer-based DETR, which is much stronger. 
Besides, our model used ResNet-50 as the backbone, resulting in relatively less gain upon QPIC.
\textbf{Second}, the three methods used different \textbf{human/object detections}.
\begin{table}[h]
    \centering
    \resizebox{0.25\textwidth}{!}{
    \begin{tabular}{c|c}
    \hline
        Methods & Object detection AP \\
        \hline
        TIN~\cite{interactiveness} & 15.38 \\
        VCL~\cite{vcl}             & 30.79 \\
        QPIC~\cite{qpic}           & 35.86 \\
        \hline
    \end{tabular}}
    \caption{Object detection performance of different methods.}
    \label{tab:od}
\end{table}
In detail, TIN used COCO pre-trained Faster-RCNN, VCL adopted HICO-DET fine-tuned Faster-RCNN, and QPIC used HICO-DET fine-tuned DETR.
The object detection performance on HICO-DET is shown in Suppl.~Tab.~\ref{tab:od}.
The better the object detection is, the higher the activity detection performance is since fewer detection errors exist in the results. Thus, less room for performance gain exists.

\subsection{HAKE-based Transfer Learning}
\begin{figure}[!ht]
    \centering
    \includegraphics[width=0.5\textwidth]{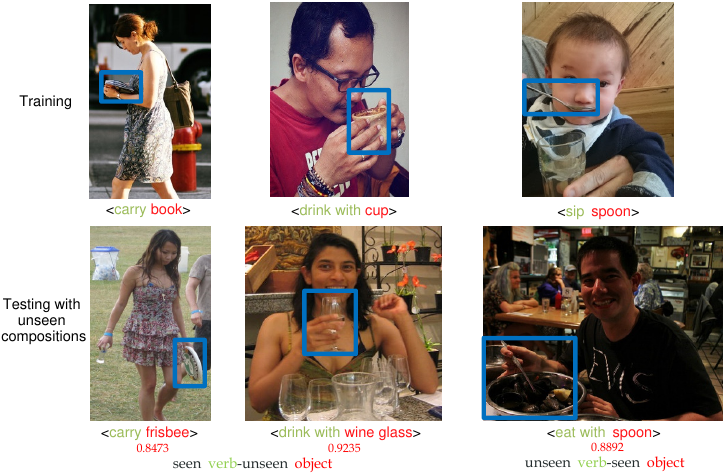}
    \caption{Visualized cases to show the generalization ability of HAKE. The first row shows the training activities, while the second row shows the testing activities that need compositional generalization ability to infer the unseen activities, \textit{i.e.}, unseen verb-object compositions. The prediction probabilities from HAKE are shown below the unseen compositions.}
    \label{fig:exp-vis}
\end{figure}

To verify the transferability of HAKE, we design transfer learning experiments on large-scale benchmarks: V-COCO~\cite{vcoco} (10,346 images, 29 activities), Ambiguous-HOI~\cite{djrn} (8,996 images, 87 activities), and AVA~\cite{ava} (seen as 234,630 images, 80 activities).
We first exclude the data from three datasets in our database and then use the rest to \textbf{pre-train} Activity2Vec and logical reasoning module with 156 activities and primitive labels. 
Then we change the classification layer (FC) size in the logical reasoning module to fit the activity categories of the target benchmark.
Finally, we freeze the primitive detector, Activity2Vec, and fine-tune the logical reasoning module only on the train set of the target dataset if it exists (Ambiguous-HOI~\cite{djrn} just contains testing data).
Here, HAKE works like the ImageNet~\cite{imagenet} and Activity2Vec is used as a pre-trained knowledge engine to promote downstream tasks.
Detailed results are reorganized and shown in Suppl.~Tab.~\ref{tab:vcoco}-\ref{tab:ava-trans}.
And we visualize some cases to show the generalization ability of HAKE in Suppl.~Fig.~\ref{fig:exp-vis}.

\begin{table}[!t]
\centering
\resizebox{0.4\textwidth}{!}{
\begin{tabular}{l c c c}
\hline
Method         &$AP_{role}(Scenario 1)$  & $AP_{role}(Scenario 2)$ \\
\hline
\hline
Gupta et al.~\cite{vcoco}                           & 31.8 & -\\
InteractNet~\cite{Gkioxari2017Detecting}           & 40.0 & -\\
GPNN~\cite{qi2018learning}                         & 44.0 & -\\
iCAN~\cite{gao2018ican}                            & 45.3 & 52.4\\
TIN~\cite{interactiveness}                         & 47.8 & 54.2 \\
\hline
iCAN~\cite{gao2018ican}+HAKE       & \textbf{49.2} ($\uparrow$8.6\%) & \textbf{55.6} ($\uparrow$6.1\%)\\ 
TIN~\cite{interactiveness}+HAKE    & \textbf{51.3} ($\uparrow$7.3\%) & \textbf{59.7} ($\uparrow$10.1\%)\\ 
\hline
\end{tabular}}
\caption{Transfer learning results on V-COCO~\cite{vcoco}. The relative improvements are shown in the brackets.}
\label{tab:vcoco}
\end{table}

\begin{table}[!t]
\centering
\resizebox{0.25\textwidth}{!}{
\begin{tabular}{l c c}
\hline
Method         & mAP \\
\hline
\hline
iCAN~\cite{gao2018ican}         & 8.14\\
TIN~\cite{interactiveness}      & 8.22\\
DJ-RN~\cite{djrn}               & 10.37 \\
\hline
TIN~\cite{interactiveness}+HAKE & \textbf{10.56} ($\uparrow$28.5\%) \\ 
DJ-RN~\cite{djrn}+HAKE & \textbf{12.68} ($\uparrow$22.3\%) \\ 
\hline
\end{tabular}}
\caption{Results comparison on Ambiguous-HOI~\cite{djrn}. The relative improvements are shown in the brackets.}
\label{tab:ambiguous-hoi}
\end{table}

\begin{table}[!t]
\begin{center}
\resizebox{0.35\textwidth}{!}{
\begin{tabular}{cc}
\hline  
Method & mAP \\
\hline  
AVA-TF~\cite{ava_tf_baseline}   & 11.4 \\
LFB-Res-50-baseline~\cite{lfb}  & 22.2 \\
LFB-Res-101-baseline~\cite{lfb} & 23.3 \\
\hline
AVA-TF~\cite{ava_tf_baseline}-HAKE   & \textbf{15.6} ($\uparrow$36.8\%) \\ 
LFB-Res-50-baseline~\cite{lfb}-HAKE  & \textbf{23.4} ($\uparrow$5.4\%) \\ 
LFB-Res-101-baseline~\cite{lfb}-HAKE & \textbf{24.3} ($\uparrow$4.3\%) \\ 
\hline
\end{tabular}}
\end{center}
\caption{Transfer learning results on image-based AVA~\cite{ava}. The relative improvements are shown in the brackets.} 
\label{tab:ava-trans}
\end{table}

\begin{table}[h]
    \centering
    \resizebox{0.25\textwidth}{!}{
    \begin{tabular}{ccc}
    \hline
    Dimension& mAP & Parameters \\
    \hline
    16 & 67.88 & 0.09M\\ 
    32 & 68.20 & 0.22M\\ 
    64 & 68.27 & 0.56M\\ 
    128 & 68.56 & 1.65M\\ 
    512 & 66.91 & 19.16M\\ 
    \hline
    \end{tabular}}
    \caption{Ablation studies on the dimension of event space. We show the model performance and parameter count for each listed dimension.}
    \label{tab:ablation_dim}
\end{table}

\subsection{Additional Ablation Study}
\noindent{\textbf{Dimension of the event space}.}
In the part of Primitive-Based Logical Reasoning, all representations are mapped into a lower-dimensional event space to accelerate the learning process. 
Here we conduct a detailed ablation study on the dimension of event space and show the results in Suppl.~Tab.~\ref{tab:ablation_dim}.
The original representations $f_{P}^L, f_{P}^V, f_{A}$ has a dimension of \textbf{2304} (three 768-dim Bert vector concatenated). 
They are first mapped into a lower-dimensional event space $e_{P}^L, e_{P}^V, e_{A}$ with MLPs due to GPU memory limitations. One single NVIDIA Titan XP GPU can afford a maximum dimension of \textbf{512}. With 1 sample per batch, the performance is \textbf{66.91} mAP. When decreasing the dimension to \textbf{128}, the performance is improved to \textbf{68.56} mAP because of \textit{increased batch size}.
The performance degrades when decreasing the dimension further, \textit{e.g.}, \textbf{67.88} mAP with a dimension of \textbf{16}. The performance loss comes from the \textit{information loss} via dimension reduction.
We finally choose the dimension of \textbf{32}. Despite a little performance loss (mAP degrades from 68.56 to \textbf{68.20}), the model parameters count is largely reduced from \textbf{1.65M} to \textbf{0.22M}, thus accelerating the learning process. Therefore, it is a near-optimal trade-off between performance and efficiency.

\noindent{\textbf{Hyper-parameters for generated rules.}}
When generating the rule candidates, there exist some hyper-parameters like 5 rules from each generator and 55 rules for each activity category. 
We give more ablation studies on these hyper-parameters.
On Ambiguous-HOI, we alter the number of generators and the number of rules per generator and evaluate the model performance. The results are listed in Suppl.~Tab.~\ref{tab:ablaton_gene}.

Originally, we equally divide $[0,1]$ into 10 intervals to obtain 11 generators with different $\beta$. 
Here, we set the number of generators as 2/6/11/21/51 respectively. As shown in Suppl.~Tab.~\ref{tab:ablaton_gene}, the model performance falls to 67.31/67.88 mAP with fewer (2/6) generators because of the decrease in rule diversity. The performance stables with 11 or more generators, \textit{i.e.}, 68.16 /68.12 for 21/51 generators, indicate that the model has fully utilized the flexibility brought by our generated rules.

Next, we modify the number of rules for each generator, \textit{i.e.}, sampling density, and we get a similar conclusion.
When sparsely sampled, the model performance falls to 67.47/67.63 mAP with 1/2 rules per generator. Thus, they are not representative enough to encode diverse primitive combinations. 
When sampled more densely, the performance improves and finally stables with 5/10/20 rules per generator. Therefore, we can make a detailed analysis of the proposed generated rules and find the most suitable hyper-parameters.

\begin{table}[ht]
    \centering
    \resizebox{0.3\textwidth}{!}{
    \begin{tabular}{cc|c}
    \hline
    \#generators & \#rules/generator & mAP \\
    \hline
    \hline
    11 & 5 & \textbf{68.20}\\
    \hline
    2 & 5 & 67.31\\
    6 & 5 & 67.88\\
    21 & 5 & 68.16\\
    51 & 5 & 68.12\\
    \hline
    11 & 1 & 67.47\\
    11 & 2 & 67.63\\
    11 & 10 & 68.15\\
    11 & 20 & 68.18\\
    \hline
    \end{tabular}}
    \caption{Ablation studies for hyper-parameters of generated rules. Experiments are done on Ambiguous-HOI under transfer learning and GT-HAKE (GT H-O boxes) setting, the same as Sec~5.4 of the main text.}
    \label{tab:ablaton_gene}
\end{table}

\section{Implementing the SCR Test}
We introduce the SCR test details in our main text as follows.

{\bf Motivation}.
To verify that HAKE has gained the knowledge to recognize the causal relations between primitives and activities. We designed the SCR test. 
We select 1,000 images from HICO-DET~\cite{hicodet} and let HAKE and human participants mask the critical regions of the activities at the same time, then we mix the images masked by humans and HAKE and allocate them to some other participants. Their task is to recognize the ongoing activities in these \textbf{masked} images. 
In this way, we can evaluate HAKE by comparing its \textit{human recognition performance degradation ability}.
If HAKE degrades the human recognition performance significantly and even approaches the degradation effect of humans' masking, we can verify that HAKE can find out the critical primitives according to the accurate primitive detection and activity reasoning. 

\begin{figure*}
	\begin{center}
		\includegraphics[width=0.8\textwidth]{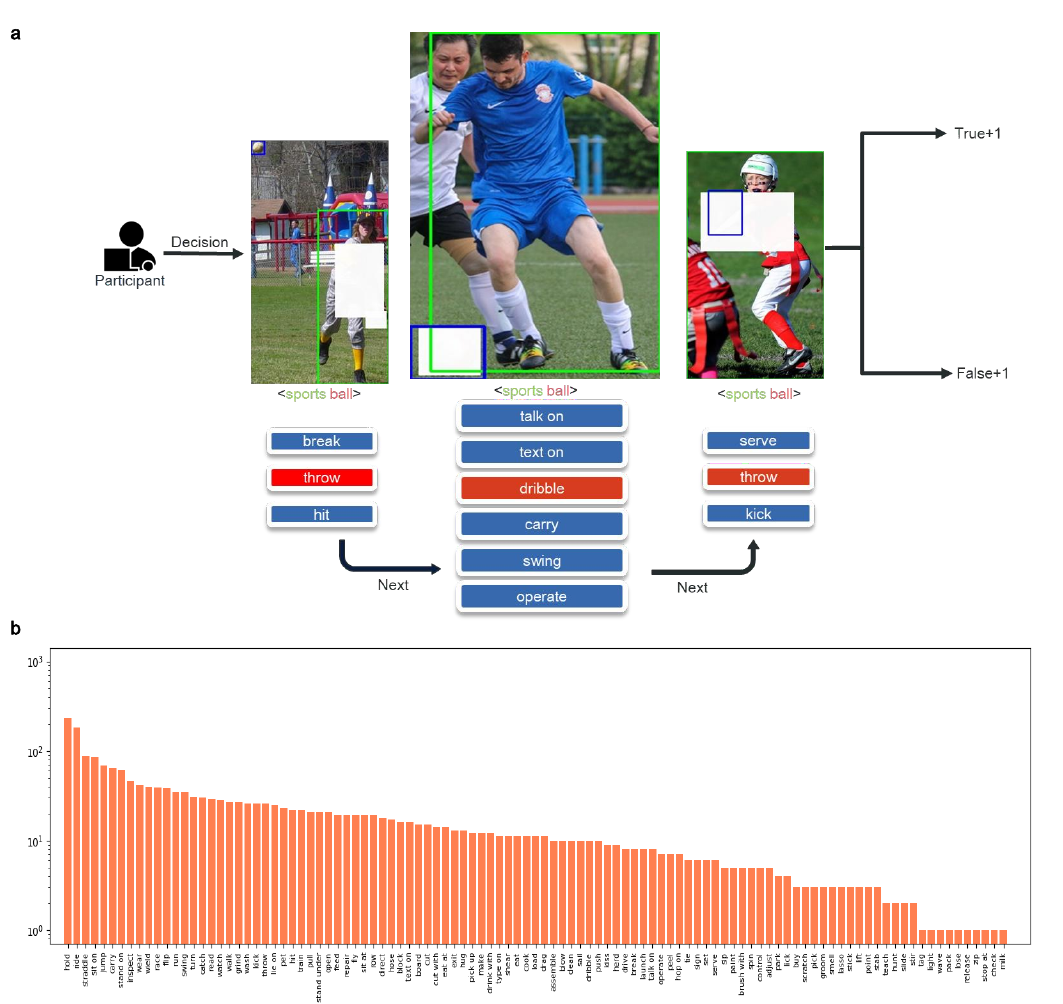}
	\end{center}
	\caption{Activity recognition process sketch and statistics of the SCR test.
	\textbf{a.} In the SCR test, to test the semantic elimination effect, each masked human-object pair (from humans/HAKE) would be given to human participants to recognize the existing activities. 
	\textbf{b.} The image numbers of different activities for the SCR test.}
	\label{figure:Details of SCR Turing test}
\end{figure*}

{\bf Data Preparation}.
Before the experiment, we first pre-process the images. We select 2,968 images from the HICO-DET~\cite{hicodet} test set to build an alternative pool. 
Several factors are considered in selecting these images:
1) All the images we select have objects that interact with a person. 
2) For clarity, the image's width and height are all greater than 200 pixels. 
3) Many other factors also affect the recognition, \textit{e.g.}, a specific body part of a person occupies all the image, crowd persons in one image, minimal persons or only a few human body parts are seen (\textit{e.g.}, only a finger pointed to something is seen, human detector and pose estimator usually fail under this kind of situation). So we first exclude these low-quality images to ensure the image quality. 
4) To avoid bias, we try our best to make the selected images cover as many activity categories as possible. 
Finally, 1,000 images are manually selected.
The image number of different activity categories is about 105, as shown in Suppl.~Fig.~\ref{figure:Details of SCR Turing test}b.

\begin{figure*}
	\begin{center}
		\includegraphics[width=0.7\textwidth]{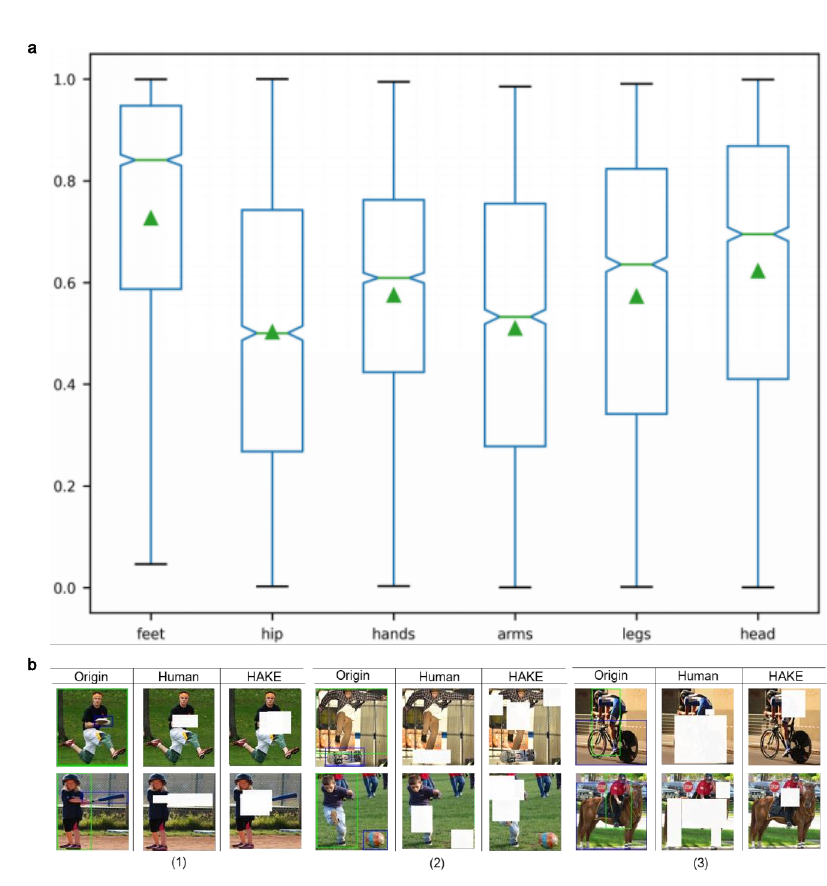}
	\end{center}
	\caption{Part level analysis. \textbf{a.} \textit{PaSta} confidences. The box plot of the estimated confidence of different body parts. The triangle represents the average confidence of each part.
	\textbf{b.} Comparison between the human and HAKE masking results. 
	\textbf{b(1).} Activities including body parts have clear semantics and direct contact with objects. 
	\textbf{b(2).} Activities including body parts indirectly contact with objects. 
	\textbf{b(3).} Activities include the persons interacting with an object and performing many different activities at the same time.}
	\label{figure:Part level analysis}
\vspace{-0.5cm}
\end{figure*}

{\bf HAKE Masking Process}.
We introduce the masking process of HAKE as follows.
For humans/objects in the images, HAKE first detects them via an object detector~\cite{faster} and a pose estimator~\cite{fang2017rmpe} to locate the whole body and body parts, including feet, legs, hands, arms, hip, and head. During the primitive detection and activity reasoning, HAKE would give confidence (0-1) to all primitives (human body parts and the interacted objects), indicating their contributions to the activity inference. More important primitives will have larger probabilities to be masked first. Thus, we rank the primitives in a confidence descending order and mask them according to the thresholds.

Notably, as different body parts usually have different characteristics, \textit{e.g.}, in daily activities, hands are usually more critical than feet. A shared threshold for all parts is unfair and would make all the masked \textit{PaSta} biased to some body parts. Thus, we calculate the confidence means and variances of different body parts of 2,968 images in Suppl.~Fig.~\ref{figure:Part level analysis}a with a box plot. Then we generate the confidence threshold of each body part according to their confidence distribution, respectively. 
Another factor is the human masking area ratio, which is a constant in large-scale masking tests. On our selected images, the human masking generates an area ratio of \textbf{0.156}. Thus, the above confidence thresholds should make the area ratio of HAKE masking close to 0.156 to keep the comparison fair. Usually, more masking regions would degrade human recognition performance more (when the masking area ratio is 1, the accuracy would be 0 because all the information of an image would be eliminated). Accordingly, we finally set the thresholds of different body parts as: 0.868 (head), 0.755 (arms), 0.763 (hands), 0.743 (hip), 0.824 (legs), 0.948 (feet).
All the selected images contain human-object interactions. Thus all the objects in them have contributions to the activity inference. However, we still have to decide when to mask these interacted objects, \textit{i.e.}, before or after some body parts. To this end, we set our policy as to when all body parts have confidences lower than their thresholds, and the object occupies a relatively small area, we only mask the object; when IoUs between the human bounding box and all body parts are all greater than 0.9, we only mask the object (even if it occupies an extensive area). 
The final mean masking area ratio of HAKE masking is \textbf{0.166}, which is very close to the human effect (\textbf{0.156}). 

{\bf Human Activity Recognition on Masked Images}.
After the above steps, we obtain 1,000 images masked by humans and 1,000 images masked by HAKE. Next, we describe the setting of human activity recognition on masked images. 
For each image, we use bounding boxes to indicate the human-object pairs. Each image can be seen as a choice question. As one person can perform multiple actions with one object simultaneously, each image can have \textbf{multiple correct answers}. 
To be close to reality, \textit{i.e.}, humans may recognize a part of the ongoing activities instead of all, we adopt the \textit{one-choice question} and set the \textbf{metric} as if \textit{anyone} of the ongoing activities is chosen, this question/image is recognized \textit{right}. To control the difficulty, we provide \textbf{60} options for each human-object pair including one of the \textit{right} right options and the other 59 \textit{wrong} options. In our test, more or fewer options would result in too low or high human performances that affect the masking effect comparison between HAKE and humans.

\begin{figure}[!ht]
\begin{center}
    \includegraphics[width=0.45\textwidth]{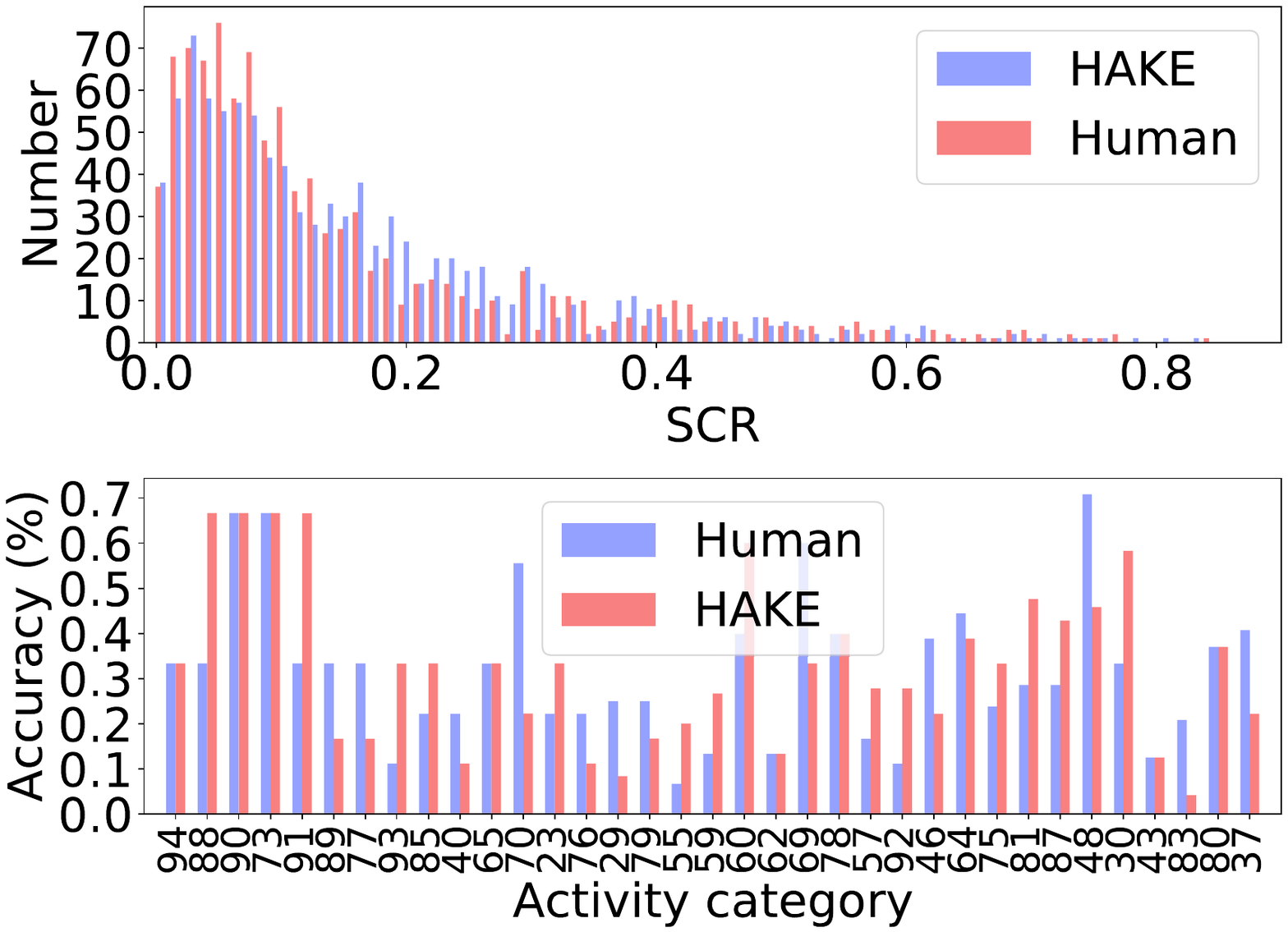}
\end{center}
	\caption{Activity recognition results of human- and HAKE- masked images. With similar masking area ratios (top), HAKE can discover key regions similar to humans and degrade human recognition performance well (bottom).}
	\label{Fig:SCR_text_scores}
\end{figure}

Next, we ask the other participants to recognize the activities from these two masked image sets and record the results.
Participants need to choose the \textit{most-likely} activity for each human-object pair, as shown in Suppl.~Fig.~\ref{figure:Details of SCR Turing test}a. 

To be fair and avoid bias, each participant will be allocated 500 randomly selected HAKE/human-masked images. During human recognition, 1,000 images are also randomly disorganized.
The final results of masked activity recognition are \textbf{35.6\% (human masking)} and \textbf{39.5\% (HAKE masking)}. 
From the result, to some extent, we can find out that HAKE achieves \textit{similar key regions localization} and \textit{recognition abilities} to humans. 
However, HAKE still has a relatively lower degradation based on a slightly higher masking area ratio.
More detailed SCRs and the human recognition accuracy of human- and HAKE- masked images are shown in Suppl.~Fig.~\ref{Fig:SCR_text_scores}.

{\bf HAKE/Human Masking Difference Analysis}.
To further study the masking difference between humans and HAKE, we conducted an additional participant recognition test. That is, we mixed the 2,000 masked images from humans and HAKE and asked the participants to make the \textbf{binary classification}: 
\textit{determine whether humans or HAKE masked each masked image}.
In total, 12 participants were involved in this test, and the classification accuracy is \textbf{59.55\%}.
In other words, HAKE can create the \textit{critical semantic region masking} very well.
Although they have similar masking abilities, the masking results of humans and HAKE still have some differences (Suppl.~Fig.~\ref{figure:Part level analysis}b): 

\textbf{1)} In Suppl.~Fig.~\ref{figure:Part level analysis}b(1), we find that both humans and HAKE can recognize the body parts directly in contact with objects well, which usually have clear semantics. 
For example, for activity \textit{hold frisbee}, both HAKE and human can quickly locate \textit{hands} and mask them. This shows that HAKE performs well on \textit{simple} activity recognition.

\textbf{2)} Meanwhile, Suppl.~Fig.~\ref{figure:Part level analysis}b(2) shows the \textit{``abstract'' ability} of HAKE. Human participants usually tend to mask the body parts \textit{directly contacting} with the object and ignore the other body parts. However, HAKE usually finds out the relationship between activity and the \textit{indirect contacted} body parts and provides prominent confidence, bringing a new perspective for us to understand the nature of activities. 
For instance, for activity \textit{jump skateboard}, human participants are prone to mask the feet directly, but HAKE takes more attention to arms, head, and hip, which are also crucial for the judgment of \textit{jump} but are often easier to be ignored.

\textbf{3)} Finally, HAKE has a certain \textit{``analytical'' ability}. As shown in Suppl.~Fig.~\ref{figure:Part level analysis}b(3), when a person is interacting with an object and performing many different activities at the same time, human participants tend to mask the object, which effectively eliminates the interaction semantics on the \textit{objectives} (\textit{bicycle}, \textit{horse}) but leaves much semantics of the \textit{verb}. For example, in the case of \textit{ride bicycle}, we may not recognize what this person rides with (\textit{e.g.}, \textit{bicycle}, \textit{motorcycle}, \textit{etc}.), but we can still easily identify the \textit{ride} after the human masking. 
Meanwhile, HAKE is quite different. It tends to mask the most related primitive it thought of. In two cases, HAKE masked the hip and saved the semantics of other activities like \textit{hold} (hands) and \textit{straddle} (legs, feet), which may be because it believes \textit{ride} is usually related to the hip most.

\section{Discussion}
Though HAKE has shown some decent abilities, it still faces a noticeable gap with human intelligence, \textit{e.g.}, it cannot always robustly perform well on all activity images.
However, its unique characteristics provide us with a new vision and show promising potential.
In general, human masking is more concise and precise visually because humans can locate the semantic entities (human/object/part) more precisely. Nevertheless, HAKE can only mask the \textbf{detected} object and human body parts which are usually not as accurate as humans' localization and contain redundant pixels. This is also why the masking area ratio of HAKE is tough to be lower than humans' effect and simultaneously keep the essential precision.
However, we believe that, with the progress of object detection, HAKE would be more precise in activity primitive discovery.

We also notice the connection between our method and syntactic approaches. 
Syntactic approaches have been applied to multiple computer vision fields, including visual grounding~\cite{shi2019visually}, image manipulation~\cite{mao2019program}, 3D shape understanding~\cite{tian2019learning, li2020multi}, and scene understanding~\cite{liu2019learning}.
More closely to our literature, Kulal et.~al.~\cite{kulal2021hierarchical} introduced the Motion Program and corresponding syntax to express motions as a composition of programs, where a program defines a certain geometric manipulation of the motion, and the syntax defines how to compose programs into a motion sequence.
This work shares some similar insights with ours. Their syntax can be seen as ``rules'' to organize the motion programs.
However, they focused more on geometric characteristics instead of semantics.
We are also inspired by the connection between syntax and our rules.
We recognize that syntax could be adopted to organize our rules more explicitly. 
Currently, we are using Boolean syntax to organize the rules by treating the primitives as Boolean variables.
Also, we also notice the potential of syntax in guiding the reasoning process.
We are optimistic about the enhancement that syntax could provide, and we leave it for future work.

\end{appendices}
\end{document}